%% file: main.tex
\documentclass[10pt,twocolumn,letterpaper]{article}

\usepackage{cvpr}              % To produce the CAMERA-READY version
\input{preamble}
\definecolor{cvprblue}{rgb}{0.21,0.49,0.74}
\usepackage[pagebackref,breaklinks,colorlinks,allcolors=cvprblue]{hyperref}
\usepackage{multirow}
\usepackage{xcolor}
\usepackage{tabularx}
\usepackage{microtype}
\def\paperID{36780} % *** Enter the Paper ID here
\def\confName{CVPR}
\def\confYear{2026}

\title{LEADER: \textcolor{red}{L}earning R\textcolor{red}{e}liable Loc\textcolor{red}{a}l-to-Global Correspon\textcolor{red}{de}nces for LiDAR \textcolor{red}{R}elocalization}

\author{Jianshi Wu$^{1,2}$ \quad Minghang Zhu$^{1,2}$ \quad Dunqiang Liu$^{1,2}$ \quad Wen Li$^{3}$ \quad Sheng Ao$^{1,2,\dagger}$ \\ Siqi Shen$^{1,2}$ \quad Chenglu Wen$^{1,2}$ \quad Cheng Wang$^{1,2}$ \\
$^{1}$Fujian Key Laboratory of Urban Intelligent Sensing and Computing\\
$^{2}$Key Laboratory of Multimedia Trusted Perception and Efficient Computing,\\
Ministry of Education of China, School of Informatics, Xiamen University, China\\
$^{3}$School of Engineering Mathematics and Technology, University of Bristol
}

\begin{document}
\maketitle
{
    \renewcommand{\thefootnote}{\fnsymbol{footnote}}
    \footnotetext[2]{Corresponding author.}
}
\input{sec/0_abstract}    
\input{sec/1_intro}
\input{sec/2_related}

\input{sec/3_method}

\input{sec/4_experiments}

\input{sec/5_ablation}
\input{sec/6_conclusion}
% {
%     \small
%     \bibliographystyle{ieeenat_fullname}
%     \bibliography{main}
% }


\input{sec/X_suppl}
{
    \small
    \bibliographystyle{ieeenat_fullname}
    \bibliography{main}
}

\end{document}

%% file: sec/0_abstract.tex
\begin{abstract}
LiDAR relocalization has attracted increasing attention as it can deliver accurate 6-DoF pose estimation in complex 3D environments. Recent learning-based regression methods offer efficient solutions by directly predicting global poses without the need for explicit map storage. However, these methods often struggle in challenging scenes due to their equal treatment of all predicted points, which is vulnerable to noise and outliers. In this paper, we propose LEADER, a robust LiDAR-based relocalization framework enhanced by a simple, yet effective geometric encoder. Specifically, a Robust Projection-based Geometric Encoder architecture which captures multi-scale geometric features is first presented to enhance descriptiveness in geometric representation. A Truncated Relative Reliability loss is then formulated to model point-wise ambiguity and mitigate the influence of unreliable predictions. Extensive experiments on the Oxford RobotCar and NCLT datasets demonstrate that LEADER outperforms state-of-the-art methods, achieving 24.1\% and 73.9\% relative reductions in position error over existing techniques, respectively. The source code is released on \url{https://github.com/JiansW/LEADER}.
\end{abstract}

%% file: sec/1_intro.tex
\section{Introduction}
\label{sec:intro}

LiDAR-based relocalization plays an important role in robotics~\cite{lu2019l3,huang2024consistency,SG-Reg}, autonomous driving~\cite{lcdnet,FMR,Difflow3d}, and virtual reality~\cite{yuan2025precision,Yu2021}. %It aims to provide global position information in environments where GNSS is unreliable or unavailable. 
Given a single LiDAR scan, LiDAR-based relocalization aims to estimate the 6 degree-of-freedom (6-DoF) pose of the sensor in the world coordinate system~\cite{lcrnet,yang2022one,lim2024quatro++}, especially in environments where GNSS is unreliable or unavailable. However, this is highly challenging due to large viewpoint variations and textureless areas.

\begin{figure}[t]
    \centering
    \includegraphics[width=0.95\columnwidth]{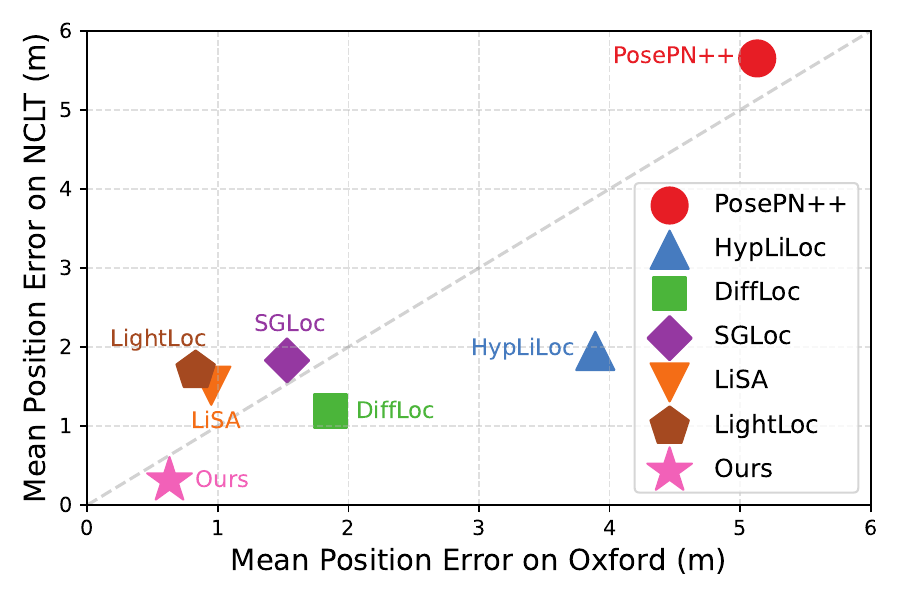}
    \caption{Mean position error comparisons on NCLT and Oxford dataset. Our method achieves superior relocalization accuracy on both datasets.}
    \label{fig:err_cmp}
\end{figure}

Currently, most methods follow the ``retrieval-then-registration'' paradigm~\cite{sc++,ring++,geotrans}, which first retrieves candidate point clouds based on feature similarity~\cite{yuan2024btc} and then estimates the 6-DoF pose through point cloud registration~\cite{spinnet,aotpami,buffer}. However, this relocalization method poses high demands on storage and communication resources, especially for city-scale maps, making it difficult to achieve efficient relocalization~\cite{locsurvey}.

Learning-based regression, which encodes scene information through neural networks, has been shown to be a promising direction for addressing these challenges~\cite{locsurvey}. These methods can be divided into two categories: Absolute Pose Regression (APR)~\cite{kendall2015posenet,wang2021pointloc,li2024diffloc} and Scene Coordinate Regression (SCR)~\cite{brachmann2017dsac,brachmann2021visual,brachmann2023accelerated,li2023sgloc,li2025lightloc}. APR directly predicts the global pose in an end-to-end fashion. In contrast, SCR estimates point correspondences between the current scan and a surrogate point cloud in world coordinates, then recovers the global 6-DoF pose via hypothesis verification methods such as RANSAC (Random Sample Consensus ~\cite{RANSAC}). Due to explicitly incorporating geometric constraints, SCR methods typically achieve superior accuracy compared with APR methods.

However, LiDAR relocalization faces two major challenges, particularly in autonomous driving scenarios: \textbf{First}, vehicles may undergo yaw rotations during driving, not always maintain a fixed orientation; \textbf{Second}, many scene elements are inherently unreliable for relocalization, as not all structures in the environment provide stable cues. As a result, current SCR methods struggle to robustly handle these challenges due to two main reasons: 1) their network architectures are not inherently robust to yaw variations, resulting in inconsistent predictions under varying viewpoints; 2) they produce erroneous correspondences against degenerate scene regions, thereby degrading relocalization performance~\cite{kendall2016modelling,huang2019prior,kfnet}.

In this paper, we propose LEADER, a simple yet effective SCR framework designed to address the aforementioned challenges in correspondence estimation. Our approach consists of two key components: 1) a robust yaw-invariant geometric encoder that generates rotation-resistant scene representations to handle yaw variations, and 2) an integrated unreliability quantification mechanism to assess point correspondence quality. In particular, we introduce a Robust Projection-based Geometric Encoder that extracts multi-scale features enhanced through projection and cyclic convolution operations, thereby strengthening representation capability and yaw invariance. This is further complemented by a Truncated Relative Reliability loss that models point-wise reliability, effectively mitigating the influence of unreliable predictions. During inference, high-reliability correspondences are used to drive RANSAC-based pose estimation. Extensive experiments demonstrate that the proposed LEADER achieves superior relocalization accuracy compared with state-of-the-art methods, as shown in \cref{fig:err_cmp}.

Overall, our contributions are three-fold:
\begin{itemize}
    \item A Robust Projection-based Geometric Encoder enhancing resilience to yaw variations in scene representation learning.
    \item A Truncated Relative Reliability loss mitigating error propagation from degenerate regions while enabling the quality estimation of point correspondences.
    \item State-of-the-art performance with 24.1\% and 73.9\% relative position error reductions on Oxford RobotCar and NCLT datasets respectively.
\end{itemize}

%% file: sec/2_related.tex
\section{Related work}
\label{sec:related}

\subsection{Traditional relocalization}
Conventional relocalization methods primarily rely on explicit representations. Retrieval-based approaches ~\cite{7052693, 7780941, 7298790, 8578568, 10378588, 9423215} identify the most similar frame from prebuilt databases, achieving fast computation at the cost of limited accuracy due to discrete pose sampling. Matching-based methods ~\cite{Wang_2019_ICCV, 10115040, 7780544, 8100137, choy20194d, 10.1109/ICRA46639.2022.9811737} formulate relocalization as point cloud registration problems by establishing correspondences between frames. However, these approaches require storing dense point cloud maps, resulting in substantial storage overhead that poses challenges for large-scale deployment.

\subsection{Absolute Pose Regression (APR)}
With advancements in hardware and deep learning, neural networks have been increasingly adopted for relocalization. Absolute Pose Regression (APR) ~\cite{wang2021pointloc, YU2022108685, wang2023hypliloc, chen2022dfnet, li2024diffloc} eliminates explicit map storage by learning implicit scene representations through end-to-end pose estimation. Pioneered by PoseNet ~\cite{kendall2016modelling} in visual relocalization, this paradigm was first adapted to LiDAR data through PointLoc ~\cite{wang2021pointloc}, which employs convolutional networks to directly regress 6-DoF poses from raw point clouds. Recent work DiffLoc ~\cite{li2024diffloc} introduces diffusion models ~\cite{10.5555/3045118.3045358, 10.5555/3295222.3295349} to refine pose predictions via iterative denoising, establishing the current state-of-the-art in APR-based LiDAR relocalization.

\subsection{Scene Coordinate Regression (SCR)}
As another implicit representation paradigm, Scene Coordinate Regression (SCR) ~\cite{li2023sgloc, yang2024lisa, brachmann2018learning6dcamera, li2025lightloc} differs from APR by decoupling coordinate prediction and pose estimation. SCR methods first regress global 3D coordinates for each scene point using learned features, and subsequently compute the optimal transformation via RANSAC-based  geometric verification. SGLoc ~\cite{li2023sgloc} introduces Scene Coordinate Regression in LiDAR-based relocalization, employs sparse convolution to extract 3D features, and employs the Attention ~\cite{8578843, liu2021densernetweaklysupervisedvisual, 9945672} mechanism for multilayer feature fusion coding, with accuracy substantially exceeding that of APR methods, while RALoc ~\cite{raloc} builds upon SGLoc to further address the rotation challenges in relocalization. The current SCR leader LiSA ~\cite{yang2024lisa} incorporates semantic ~\cite{10203552} priors to differentiate point-wise contributions, demonstrating varying contributions of scene points on final relocalization accuracy.

\begin{figure*}[t]
    \centering
    \includegraphics[width=0.9\linewidth]{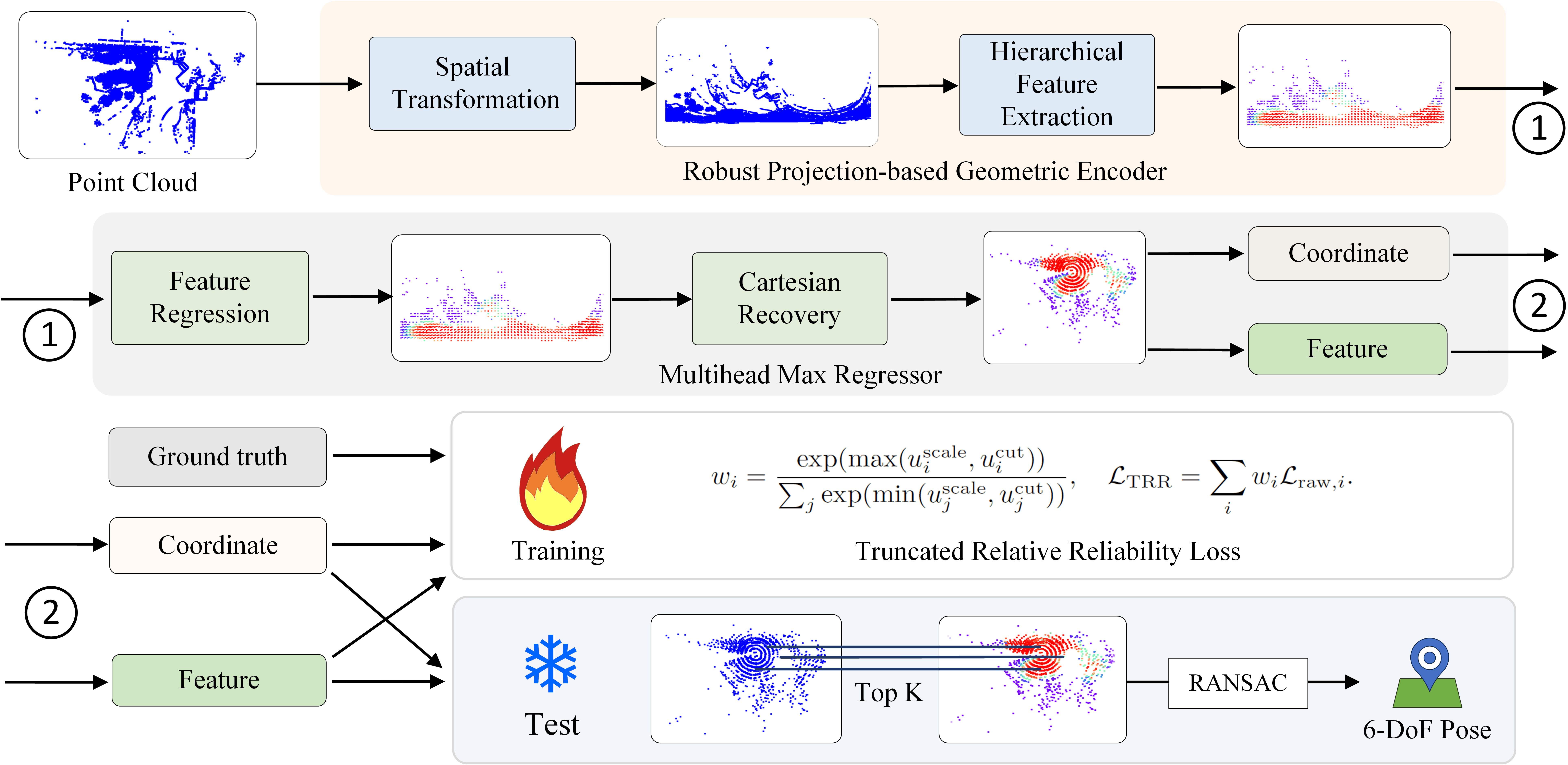}
    % \rule{0.98\textwidth}{0.45\textwidth}
    \caption{The pipeline of the proposed LEADER. Raw point clouds undergo spatial transformation to establish yaw-invariant spatial representations. Hierarchical feature extraction then derives multi-level fused features by cyclic convolution. The Regressor processes features from the Robust Projection-based Geometric Encoder (RPGE) to output predict coordinates and reliability values, followed by Cartesian Recovery. During training, the Truncated Relative Reliability (TRR) loss optimizes predictions against ground truth. During inference, high-reliability points are selected via TRR filtering for robust 6-DoF pose estimation through RANSAC-based estimator.}
    \label{fig:method}
\end{figure*}

%% file: sec/3_method.tex
\section{Method}
\label{sec:method}

The proposed framework, LEADER, takes a single LiDAR scan as input and directly outputs a global 6-DoF pose. As shown in \cref{fig:method}, our network is trained end-to-end using the Truncated Relative Reliability loss. During inference, high-reliability correspondences are selected and refined via a RANSAC-based solver to recover the final pose.
%The methodological pipeline of LEADER is illustrated in Fig.~\ref{fig:method}.

\subsection{Robust Projection-based Geometric Encoder}
\label{sec:robust_encoder}

Our encoder architecture transforms raw point clouds into geometrically enhanced representations resilient to viewpoint variations, particularly yaw rotations. The processing pipeline consists of two main stages: Spatial Transformation and Hierarchical Feature Extraction, which work collaboratively to achieve yaw-robust representation learning.

\textbf{Spatial Transformation:}
Given a raw point cloud $\mathcal{P}_{\text{raw}} = \{\mathbf{p}_i = (x_i, y_i, z_i)\}_{i=1}^N$ and transformation $\mathbf{T}$, we first estimate its ground plane using Patchwork++ ~\cite{lee2022patchworkpp}. The point cloud is then rectified to the horizontal plane following ~\cite{DBLP:conf/rss/SegalHT09}, yielding the planar rectification matrix $\mathbf{T}_{\text{plane}}$ and transformed point cloud $\mathcal{P}' = \{\mathbf{p}_i' = (x_i', y_i', z_i')\}_{i=1}^N$.

Subsequently, we apply a geometrically constrained cylindrical projection inspired by Scan Context ~\cite{pan2024tro, 8593953} and Mercator mapping. This transformation converts rectified points into cylindrical coordinates through:
\begin{equation}
\begin{aligned}
x^\mathbf{p} &= s \cdot \arctan2(y', x') \\
y^\mathbf{p} &= \sqrt{x'^2 + y'^2} \\
z^\mathbf{p} &= z' 
\end{aligned} \quad \Rightarrow \quad \mathbf{p}_i^\mathbf{p} = (x_i^\mathbf{p}, y_i^\mathbf{p}, z_i^\mathbf{p}),
\label{eq:geo_proj}
\end{equation}
where scaling factor $s$ controls angular resolution. The projected points $\mathbf{p}_i^\mathbf{p}$ are treated as Cartesian coordinates for subsequent feature extraction. Voxelization with cell size $\delta$ is then applied to generate a structural representation $\mathbf{v}$. Each voxel is denoted as $\mathbf{v}_{x, y, z}=(x^\mathbf{v},y^\mathbf{v},z^\mathbf{v})$, with circumferential resolution of $\text{L}_x = 2\pi s/\delta$. The final point cloud $\mathcal{P}^\mathbf{v}$ retains one point per voxel, establishing a consistent spatial structure for robust feature learning.

\textbf{Hierarchical Feature Extraction:}
We transform the spatially transformed point cloud $\mathcal{P}^\mathbf{v}$ into geometrically robust feature through a multi-scale processing pipeline. To maintain yaw-invariant representation, we construct the initial feature vector for each point $\mathbf{p}_i^\mathbf{v}$ as $\mathbf{f}_i = [y_{i}^\mathbf{v}, z_{i}^\mathbf{v}, I_i^\mathbf{v}] \in \mathbb{R}^3$, with $I_i^\mathbf{v}$ representing intensity. These features are intentionally selected for their independence from yaw variations, thereby preserving rotational invariance in the initial representation.

To address the spatial continuity issue at the yaw boundaries introduced by cylindrical projection, we employ cyclic sparse convolution with symmetric padding. Specifically, we generate an expanded point set through:
\begin{equation}
\begin{aligned}
\mathcal{P}_{\text{pad}}^\mathbf{v} = &\mathcal{P}^\mathbf{v} \cup \left\{\mathbf{p}_i^\mathbf{v} + [2\pi s, 0, 0] \mid x_{i}^\mathbf{v} < w\right\} \\
& \cup \left\{\mathbf{p}_i^\mathbf{v} - [2\pi s, 0, 0] \mid x_{i}^\mathbf{v} > \text{L}_x - w\right\},
\label{eq:cyclic_padding}
\end{aligned}
\end{equation}
where $w$ represents the convolution kernel width. After convolutional processing, points outside the original range $[0, \text{L}_x]$ are discarded, ensuring seamless feature transition across the yaw boundary while maintaining structural continuity.

We design a U-Net-style ~\cite{10.1007/978-3-319-24574-4_28, qi2017pointnetplusplus} architecture for multi-scale feature fusion. The network follows a UNet-style architecture with five downsampling stages and a single upsampling stage applied after the fifth downsampling. Each convolutional layer is implemented as a cyclic sparse convolution to consistently handle cylindrical projection effects. Channel dimensions progressively expand along the downsampling path as [32,64,128,256,384]. After upsampling from the fifth stage, the resulting features are concatenated with those from the fourth downsampling stage and projected via a fully connected layer to 512 dimensions, yielding enriched embeddings that capture both local geometric details and global contextual information. Each downsampling and fusion step is followed by two 3×3 convolutional layers.

\subsection{Multihead Max Regressor}

This module transforms encoded 512-dimensional features into global scene coordinates while recovering Cartesian representations, ensuring geometrically consistent outputs for subsequent pose estimation. The regression process consists of two sequential stages: multi-layer feature regression and Cartesian coordinate recovery.

\textbf{Feature Regression:}
Given input features $\mathbf{F} \in \mathbb{R}^{N}$ ($N=512$) from the encoder, we employ a multi-head projection mechanism defined as:
\begin{equation}
\mathbf{F}' = \mathop{\mathrm{\max}}_{k} \left( \mathrm{Reshape}_{k \times N}\left( \mathbf{W}_d \mathbf{F} + \mathbf{b}_d \right) \right),
\label{eq:mhms}
\end{equation}
where $\mathbf{W}_d \in \mathbb{R}^{kN \times N}$ projects features to $kN$ dimensions ($k=4$), followed by reshaping into $k$ parallel $N$-dimensional heads. The $\max(\cdot)$ operation selects the strongest activation per dimension across heads. The regression backbone consists of $l=5$ stacked layers implementing \cref{eq:mhms}, each followed by LayerNorm and LeakyReLU activation, concluding with a final fully connected layer to output a 4-D features that includes 3-D of scene coordinates $\mathbf{\hat{c}}_i = (x_i^\mathrm{w}, y_i^\mathrm{w}, z_i^\mathrm{w})$ in the world frame and 1-D of reliability scores $u_i$, which are essential for the subsequent Truncated Relative Reliability (TRR) loss.

\textbf{Cartesian Recovery:}
To reconstruct original Cartesian coordinates from regressed geometric representations, we invert the projection defined in Eq.~\eqref{eq:geo_proj}. Continuous coordinates are derived from the voxel centers $\mathbf{q}_{x,y,z}=(x^\mathbf{q},y^\mathbf{q},z^\mathbf{q})$ output by the RPGE:
\begin{align}
\hat{x} &= y^\mathbf{q} \cos\left(\frac{x^\mathbf{q}}{s}\right), &
\hat{y} &= y^\mathbf{q} \sin\left(\frac{x^\mathbf{q}}{s}\right), & 
\hat{z} &= z^\mathbf{q}.
\label{eq:inv_proj}
\end{align}
The final scene coordinates $\mathcal{\hat{P}} = \{(\hat{x}_i, \hat{y}_i, \hat{z}_i)\}_{i=1}^M$ in Cartesian space preserve the yaw-robust geometric consistency established by the encoder, thereby completing the transformation from encoded features to geometrically stable world coordinateds.

\begin{figure}[t]
    % \hfill
    \hspace{0.03\linewidth}
    \centering
    \begin{subfigure}{0.42\linewidth}
        \centering
        {\color{gray}\fbox{\includegraphics[width=0.9\linewidth]{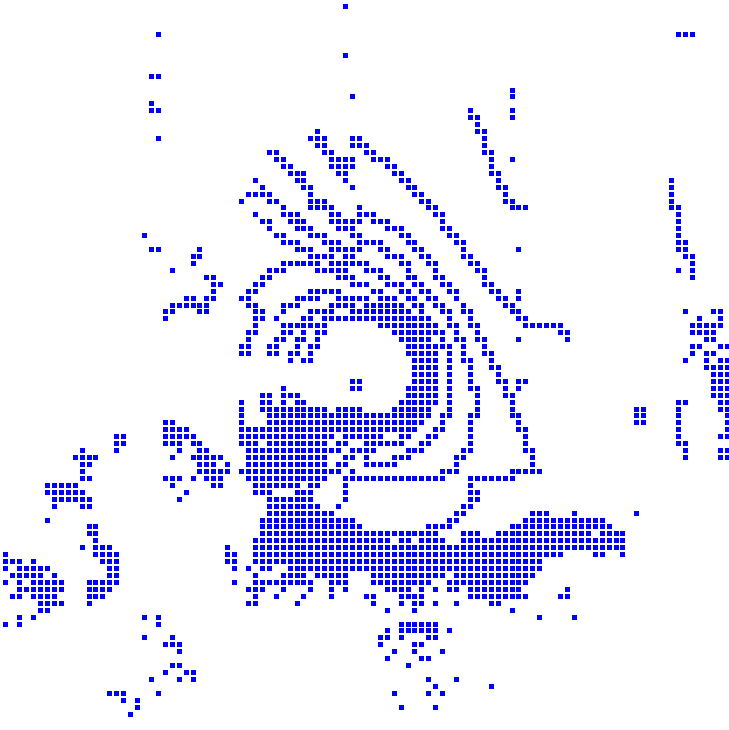}}}
        \label{fig:pointcloud_normal}
    \end{subfigure}
    \hfill
    \begin{subfigure}{0.42\linewidth}
        \centering
        {\color{gray}\fbox{\includegraphics[width=0.9\linewidth]{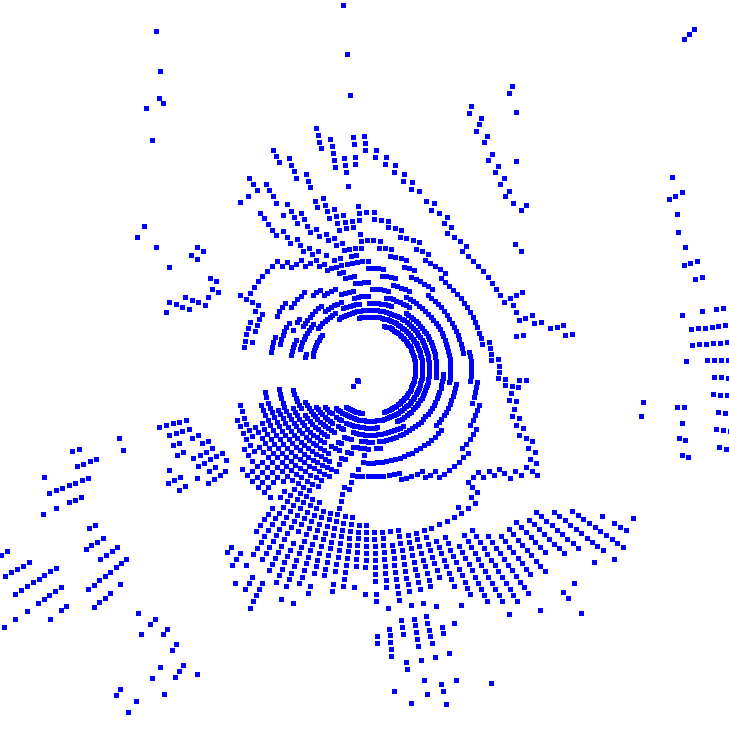}}}
        \label{fig:pointcloud_project}
    \end{subfigure}
    \hspace{0.03\linewidth}
    
    \vspace{0.001\textwidth}

    \caption{Comparison of point cloud density distribution. \textbf{Left:} The point cloud is directly voxelized. \textbf{Right:} The point cloud is processed through the Spatial Transformation, voxelized, and then recovered via Cartesian Recovery.}
    \label{fig:project_comparation}
\end{figure}

As shown in \cref{fig:project_comparation}, distinct density distributions are observed between the point clouds processed via different pipelines. The point cloud that is directly voxelized without any transformation demonstrates a relatively uniform point density. In contrast, the point cloud on processed through the Spatial Transformation followed by voxelization and Cartesian Recovery, exhibits a clear variation in density, with points appearing denser in nearby regions and sparser in areas farther away.

\subsection{Loss function} 
\label{sec:lossfn}

We propose the \textbf{Truncated Relative Reliability (TRR)} loss to optimize the estimation network and mitigate error propagation from geometrically ambiguous points. Given the predicted scene coordinates $\mathbf{\hat{c}}_i$ and reliability scores $u_i$ from the regressor, the TRR loss operates through three principled components:

\textbf{1) Geometric error calculation:} The ground truth scene coordinates $\mathbf{c}_i^\text{gt}$ are derived from the point cloud $\mathcal{\hat{P}}$ and its corresponding transformation matrix T. The Euclidean distance between predicted and ground truth coordinated defines the raw geometric loss:
\begin{equation}
\mathcal{L}_{\text{raw},i} = \|\mathbf{c}_i^{\text{gt}} - \mathbf{\hat{c}}_i\|_2.
\label{eq:rawloss}
\end{equation}

\textbf{2) Reliability calibration:} To prevent gradient vanishing while bounding score magnitudes, we employ arc-tangent scaling with clamping:
\begin{equation}
u_i^{\text{scale}} = \mathop{\mathrm{arctan}}(u_i) \cdot K_s,
\label{eq:scale}
\end{equation}
\begin{equation}
u_i^{\text{cut}} = \arctan(\text{clamp}(u_i, -10\pi, 10\pi)) \cdot K_s.
\label{eq:trunc}
\end{equation}
where $K_s = \ln 10/\pi$ is a normalization constant to bound output magnitudes.

\textbf{3) Gradient-aware truncation:} The final loss integrates these components through normalized exponential weighting:
\begin{equation}
w_i = \frac{\exp(\max(u_i^{\text{scale}},u_i^{\text{cut}}))}{\sum_j \exp(\min(u_j^{\text{scale}}, u_j^{\text{cut}}))}, 
\label{eq:point_weight}
\end{equation}
\begin{equation}
\mathcal{L}_{\text{TRR}} = \sum_i w_i \mathcal{L}_{\text{raw},i}.
\label{eq:finaloss}
\end{equation}

\textbf{Design rationale:} The loss \cref{eq:finaloss} encourages competition among points: the network learns to suppress reliability for hard or ambiguous samples (reducing $w_i$) while enhancing it for discriminative points, thereby focusing model capacity on more reliable features. This self-rebalancing mechanism allows the model to autonomously allocate learning resources, sacrificing precision on low-quality regions (\eg, weak textures or dynamic objects) to improve overall representation quality.

\begin{table*}[t]
    \centering
    \small
    \caption{\textbf{Mean position error (m) and mean orientation error (\textdegree) on the Quality-enhanced Oxford dataset}. All values are given as (m, \textdegree). Lower values are better, with the best highlighted using bold and the second-best highlighted using underline. In both errors, LEADER achieved the best or second-best performance among all baselines.}
    \setlength{\tabcolsep}{5mm}
    \begin{tabular}{c|c|cccc|c}
        \toprule
        \multicolumn{2}{c|}{Method} & 15-13-06-37 & 17-13-26-39 & 17-14-03-00 & 18-14-14-42 & Average \\
        \midrule
        \multirow{4}{*}{APR} & PointLoc ~\cite{wang2021pointloc} & 10.75, 2.36 & 11.07, 2.21 & 11.53, 1.92 & 9.82, 2.07 & 10.79, 2.14 \\
         & PosePN++ ~\cite{YU2022108685} & 4.54, 1.83 & 6.44, 1.78 & 4.89, 1.55 & 4.64, 1.61 & 5.13, 1.69 \\
         & HypLiLoc ~\cite{wang2023hypliloc} & 5.03, 1.46 & 4.31, 1.43 & 3.61, 1.11 & 2.61, \underline{1.09} & 3.89, 1.27 \\
         & DiffLoc ~\cite{li2024diffloc} & 2.03, \textbf{1.04} & 1.78, \textbf{0.79} & 2.05, \textbf{0.83} & 1.56, \textbf{0.83} & 1.86, \textbf{0.87} \\
        % \cmidrule{1-2}
        \cline{1-2}
        \multirow{4}{*}{SCR} & SGLoc ~\cite{li2023sgloc} & 1.79, 1.67 & 1.81, 1.76 & 1.33, 1.59 & 1.19, 1.39 & 1.53, 1.60 \\
         & LiSA ~\cite{yang2024lisa} & 0.94, 1.10 & 1.17, 1.21 & 0.84, 1.15 & 0.85, 1.11 & 0.95, 1.14 \\
         & LightLoc ~\cite{li2025lightloc} & \underline{0.82}, 1.12 & \underline{0.85}, \underline{1.07} & \underline{0.81}, 1.11 & \underline{0.82}, 1.16 & \underline{0.83}, 1.12 \\
         & RALoc ~\cite{raloc} & 1.52, 1.28 & 1.71, 1.27 & 1.37, 1.20 & 1.41, 1.23 & 1.51, 1.24 \\
         & \textbf{Ours} & \textbf{0.69}, \underline{1.08} & \textbf{0.64}, 1.16 & \textbf{0.59}, \underline{1.10} & \textbf{0.62}, \underline{1.09} & \textbf{0.63}, \underline{1.11} \\
        \bottomrule
        % \Xhline{1.0pt}
    \end{tabular}
    \label{tab:oxford_results}
\end{table*}

\begin{figure*}[t]
    \centering
    % \hspace{0.05\textwidth}
    \begin{subfigure}{0.24\linewidth}
        \includegraphics[width=1\linewidth]{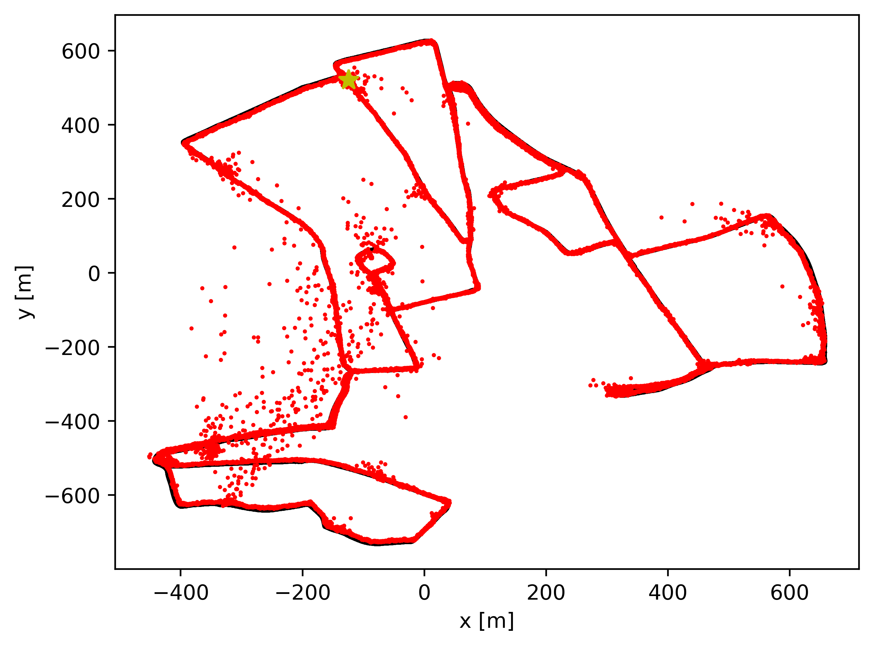}
       % \rule{1.0\linewidth}{.7\linewidth}
        \caption{PosePN++}
        \label{fig:oxford_posepnpp}
    \end{subfigure}
    \hfill
    \begin{subfigure}{0.24\linewidth}
        \includegraphics[width=1\linewidth]{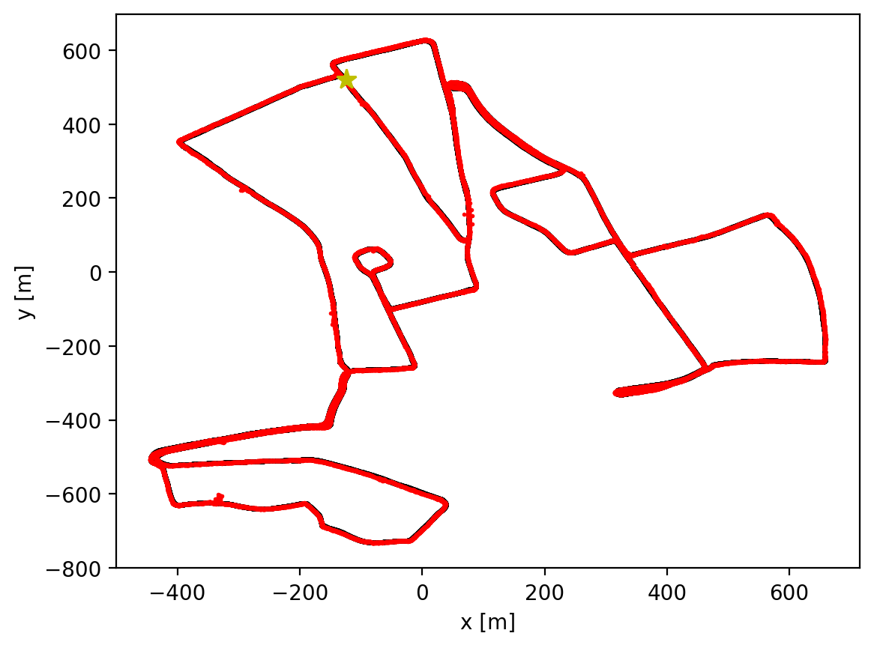}
       % \rule{1.0\linewidth}{.7\linewidth}
        \caption{DiffLoc}
        \label{fig:oxford_diffloc}
    \end{subfigure}
    \hfill
    \begin{subfigure}{0.24\linewidth}
        \includegraphics[width=1\linewidth]{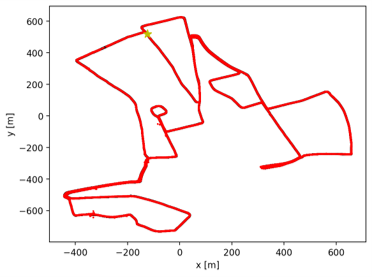}
       % \rule{1.0\linewidth}{.7\linewidth}
        \caption{LightLoc}
        \label{fig:oxford_lightloc}
    \end{subfigure}
    % \hspace{0.05\textwidth}
    \hfill
    \begin{subfigure}{0.24\linewidth}
        \includegraphics[width=1\linewidth]{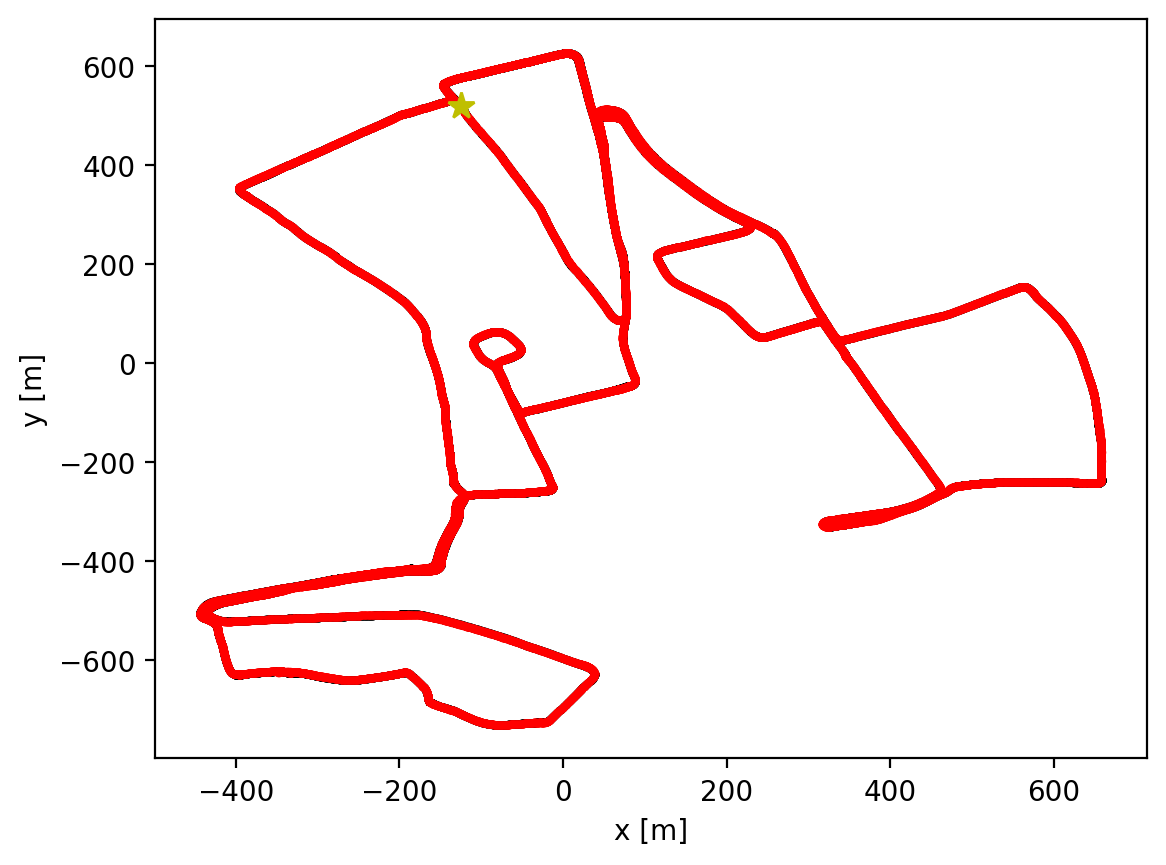}        
        \caption{LEADER}
        \label{fig_oxford_LEADER}
    \end{subfigure}
    
    \vspace{0.001\textwidth}

    % \hspace{0.05\textwidth}
    % \begin{subfigure}{0.25\linewidth}
    %     \includegraphics[width=1\linewidth]{../diagram/qeoxford2639_lisa.png}
    %    % \rule{1.0\linewidth}{.7\linewidth}
    %     \caption{LiSA}
    %     \label{fig:lisa}
    % \end{subfigure}
    % \hfill
    % \begin{subfigure}{0.25\linewidth}
    %     \includegraphics[width=1\linewidth]{../diagram/qeoxford2639_lightloc.png}
    %     % \rule{1.0\linewidth}{.7\linewidth}
    %     \caption{LightLoc}
    %     \label{fig:LightLoc}
    % \end{subfigure}
    % \hfill
    % \begin{subfigure}{0.25\linewidth}
    %     \includegraphics[width=1\linewidth]{../diagram/qeoxford2639_LEADER.png}
    %    % \rule{1.0\linewidth}{.7\linewidth}
    %     \caption{LEADER}
    %     \label{fig:LEADER}
    % \end{subfigure}
    % \hspace{0.05\textwidth}
    % \vspace{0.001\textwidth}
    \caption{Visualization results of part of the methods on trajectory 17-13-26-39 in Quality-enhanced Oxford dataset. The black and red points represent the true and predicted results, respectively. The star indicates the starting point.}
    \label{fig:oxford_view_traj}
\end{figure*}

\subsection{Inference}

During inference, the network predicts both scene coordinates $\mathbf{\hat{c}}_i = (x_i^\mathrm{w}, y_i^\mathrm{w}, z_i^\mathrm{w})$ and reliability scores $u_i$ for each point. We establish 3D-3D correspondences through adaptive reliability thresholding:
\begin{equation}
\mathcal{S} = \begin{cases}
\text{argtopk}(u_i) & \text{if } |\text{argtopk}(u_i)| \geq 50, \\
\{1,\ldots,N\} & \text{otherwise}.
\end{cases}
\label{eq:selection}
\end{equation}
where $\text{argtopk}(\cdot)$ selects indices of points with top reliability scores. Let $\mathcal{P}^\mathrm{l} = \{\mathbf{p}_i^\mathrm{l} = (x_i^\mathrm{l}, y_i^\mathrm{l}, z_i^\mathrm{l})\}_{i\in\mathcal{S}}$ denote selected local coordinates and $\mathcal{P}^{\text{pred}} = \{\mathbf{\hat{c}}_i\}_{i\in\mathcal{S}}$ their predicted global counterparts. The 6-DoF pose $\mathbf{T^*} \in \mathrm{SE}(3)$ is estimated via ~\cite{Chen_2022_CVPR}:
\begin{equation}
\mathbf{T^*} = \underset{\mathbf{T}}{\arg\min} \rho \left(\sum_{i\in\mathcal{S}} \|\mathbf{T}\cdot\mathbf{p}_i^\mathrm{l} - \mathbf{\hat{c}}_i\|_2\right),
\label{eq:ransac}
\end{equation}
where $\rho(\cdot)$ is the estimator. 

Since the ground plane transformation $\mathbf{T}_{\text{plane}}$ is applied, the final global pose requires inverse compensation:
\begin{equation}
\mathbf{T_{\text{final}}} = \mathbf{T^*} \cdot \mathbf{T_{\text{plane}}^{-1}}.
\label{eq:pose_comp}
\end{equation}
This two-stage approach decouples learning-based correspondence prediction from geometric verification. The compensation step in \cref{eq:pose_comp} ensures consistency between the rectified coordinates and original global frame.

%% file: sec/4_experiments.tex
\begin{table*}[t]
    \centering
    \small
    \caption{\textbf{Mean position error (m) and mean orientation error (\textdegree) on the NCLT dataset}. All values are given as (m, \textdegree). Lower values are better, with the best highlighted using bold and the second-best highlighted using underline. In both errors, LEADER achieved the best performance among all baselines.}
    \setlength{\tabcolsep}{5mm}
    \begin{tabular}{c|c|cccc|c}
        \toprule
        \multicolumn{2}{c|}{Method} & 2012-02-12 & 2012-02-19 & 2012-03-31 & 2012-05-26 & Average \\
        \midrule
        \multirow{4}{*}{APR} & PointLoc ~\cite{wang2021pointloc} & 7.23, 4.88 & 6.31, 3.89 & 6.71, 4.32 & 10.02, 5.32 & 7.57, 4.60 \\
        & PosePN++ ~\cite{YU2022108685} & 4.97, 3.75 & 3.68, 2.65 & 4.35, 3.38 & 9.59, 4.49 & 5.65, 3.57 \\
        & HypLiLoc ~\cite{wang2023hypliloc} & 1.71, 3.56 & 1.68, 2.69 & 1.52, 2.90 & 2.90, 3.47 & 1.95, 3.16 \\
        & DiffLoc ~\cite{li2024diffloc} & 0.99, 2.40 & 0.92, 2.14 & 0.98, 2.27 & \underline{1.88}, \underline{2.43} & \underline{1.19}, \underline{2.31} \\
        \cline{1-2}
        \multirow{4}{*}{SCR} & SGLoc ~\cite{li2023sgloc} & 1.20, 3.08 & 1.20, 3.05 & 1.12, 3.28 & 3.81, 4.74 & 1.83, 3.54 \\
        & LiSA ~\cite{yang2024lisa} & \underline{0.97}, \underline{2.23} & \underline{0.91}, \underline{2.09} & 0.87, \underline{2.21} & 3.30, 2.84 & 1.51, 2.34 \\
        & LightLoc ~\cite{li2025lightloc} & 1.03, 2.91 & 0.92, 2.63 & \underline{0.86}, 2.78 & 3.99, 3.49 & 1.70, 2.95 \\
        & RALoc ~\cite{raloc} & 1.61, 4.71 & 1.61, 4.88 & 1.58, 4.71 & 3.51, 5.40 & 2.07, 4.92 \\
        & \textbf{Ours} & \textbf{0.37}, \textbf{1.89} & \textbf{0.27}, \textbf{1.75} & \textbf{0.29}, \textbf{1.81} & \textbf{0.32}, \textbf{1.79} & \textbf{0.31}, \textbf{1.81} \\
        \bottomrule
        \end{tabular}
    \label{tab:nclt_results}
\end{table*}

\begin{figure*}
    \centering
    \begin{subfigure}{0.24\linewidth}
        \includegraphics[width=1\linewidth]{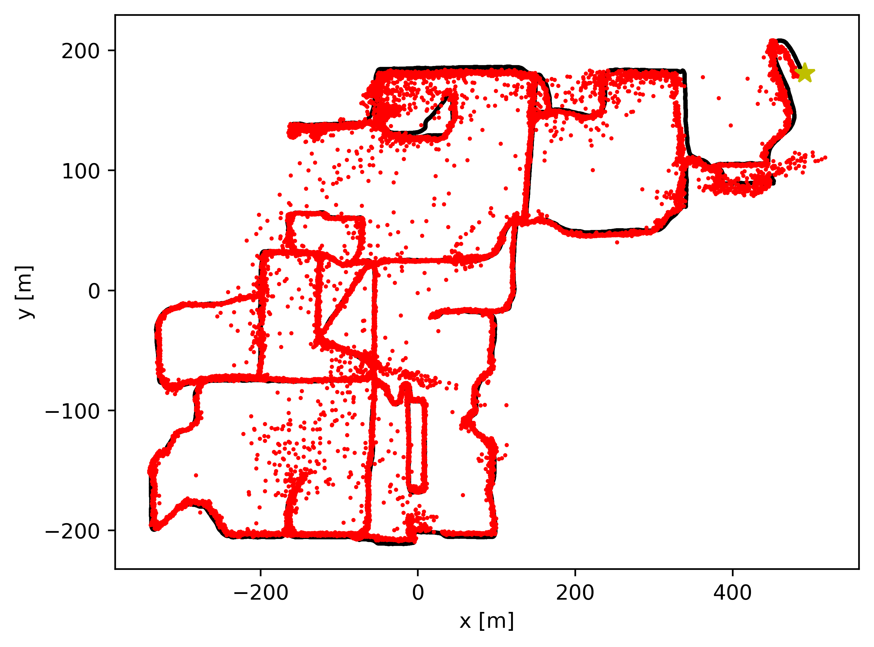}
       % \rule{1.0\linewidth}{.7\linewidth}
        \caption{PosePN++}
        \label{fig:nclt_posepnpp}
    \end{subfigure}
    \hfill
    \begin{subfigure}{0.24\linewidth}
        \includegraphics[width=1\linewidth]{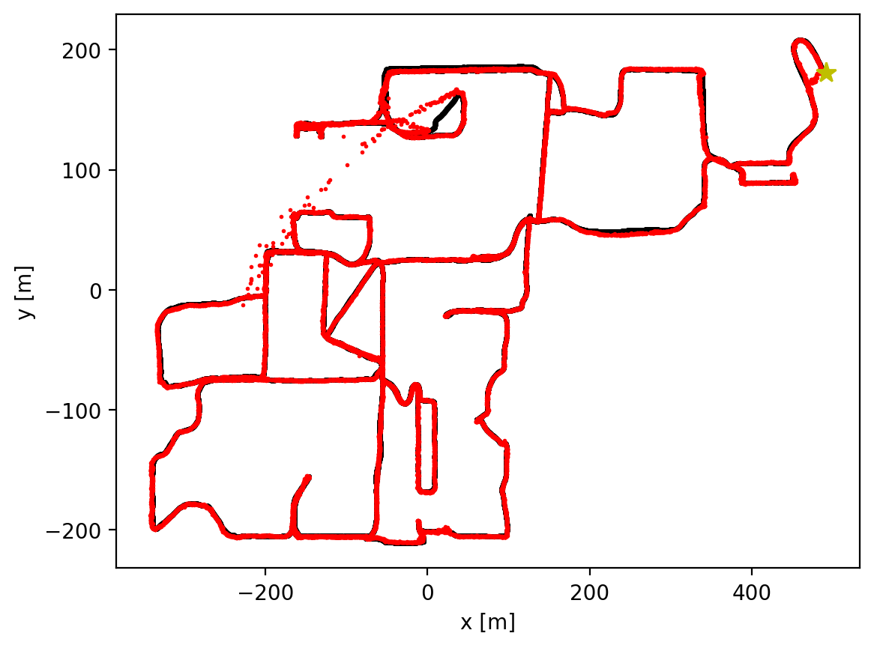}
       % \rule{1.0\linewidth}{.7\linewidth}
        \caption{DiffLoc}
        \label{fig:nclt_diffloc}
    \end{subfigure}
    \hfill
    \begin{subfigure}{0.24\linewidth}
        \includegraphics[width=1\linewidth]{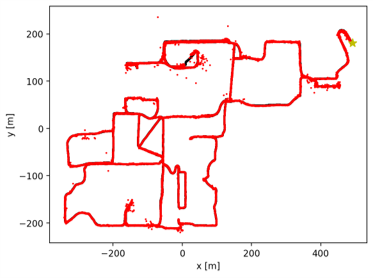}
       % \rule{1.0\linewidth}{.7\linewidth}
        \caption{LightLoc}
        \label{fig:nclt_lightloc}
    \end{subfigure}
    \hfill
    \begin{subfigure}{0.24\linewidth}
        \includegraphics[width=1\linewidth]{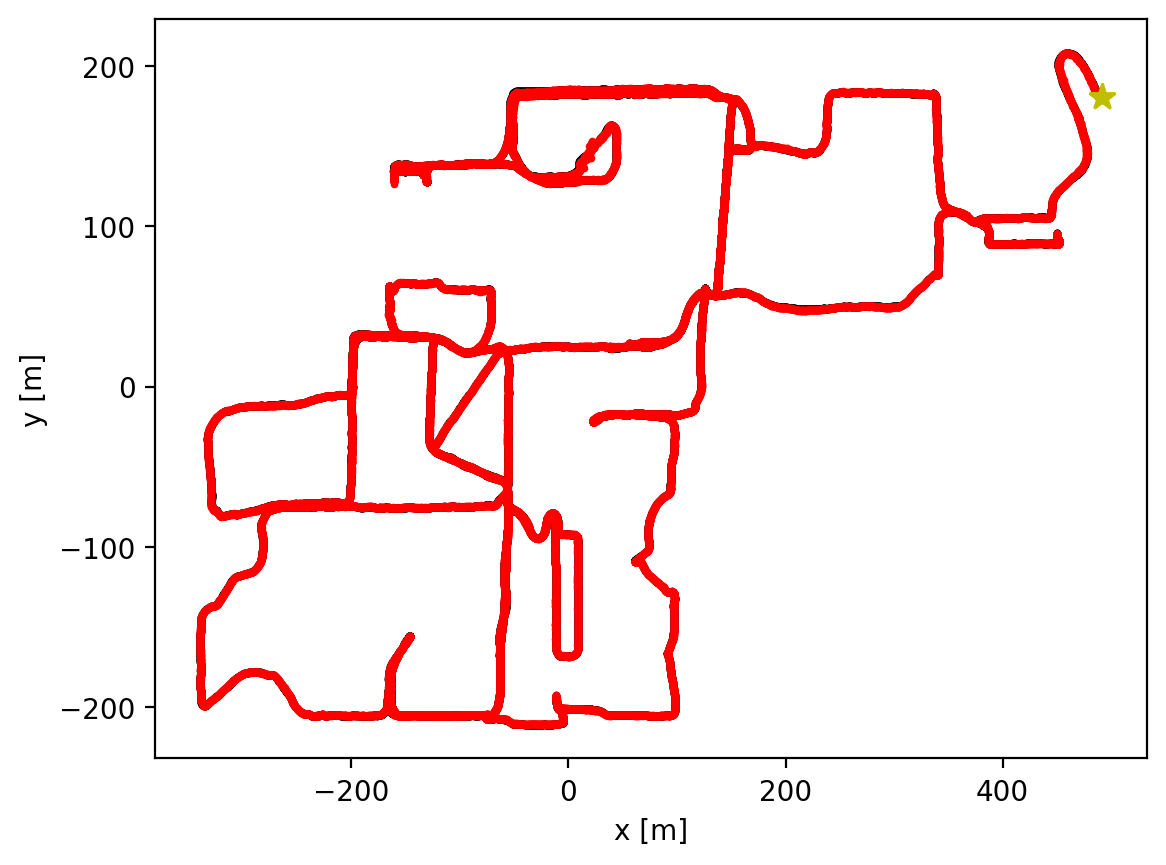}
        \caption{LEADER}
        \label{fig:nclt_LEADER}
    \end{subfigure}
    \vspace{0.001\textwidth}
    % \hspace{0.05\textwidth}
    % \begin{subfigure}{0.25\linewidth}
    %     \includegraphics[width=1\linewidth]{../diagram/nclt0526_hypliloc.png}
    %    % \rule{1.0\linewidth}{.7\linewidth}
    %     \caption{HypLiLoc}
    %     \label{fig:hypliloc}
    % \end{subfigure}
    % \hfill
    % \begin{subfigure}{0.25\linewidth}
    %     \includegraphics[width=1\linewidth]{../diagram/nclt0526_diffloc.png}
    %    % \rule{1.0\linewidth}{.7\linewidth}
    %     \caption{DiffLoc}
    %     \label{fig:diffloc}
    % \end{subfigure}
    % \hfill
    % \begin{subfigure}{0.25\linewidth}
    %     \includegraphics[width=1\linewidth]{../diagram/nclt0526_sgloc.png}
    %    % \rule{1.0\linewidth}{.7\linewidth}
    %     \caption{SGLoc}
    %     \label{fig:sgloc}
    % \end{subfigure}
    % \hspace{0.05\textwidth}

    % \vspace{0.001\textwidth}
    
    % \hspace{0.05\textwidth}
    % \begin{subfigure}{0.25\linewidth}
    %     \includegraphics[width=1\linewidth]{../diagram/nclt0526_lisa.png}
    %    % \rule{1.0\linewidth}{.7\linewidth}
    %     \caption{LiSA}
    %     \label{fig:lisa}
    % \end{subfigure}
    % \hfill
    % \begin{subfigure}{0.25\linewidth}
    %     \includegraphics[width=1\linewidth]{../diagram/nclt0526_lightloc.png}
    %    % \rule{1.0\linewidth}{.7\linewidth}
    %     \caption{LightLoc}
    %     \label{fig:lightloc}
    % \end{subfigure}
    % \hfill
    % \begin{subfigure}{0.25\linewidth}
    %     \includegraphics[width=1\linewidth]{../diagram/nclt0526_LEADER.png}
    %    % \rule{1.0\linewidth}{.7\linewidth}
    %     \caption{LEADER}
    %     \label{fig:LEADER}
    % \end{subfigure}
    % \hspace{0.05\textwidth}
    % \vspace{0.001\textwidth}
    \caption{Visualization results of part of the methods on trajectory 2012-05-26. The black and red points represent the true and predicted results, respectively. The star indicates the starting point.}
    \label{fig:nclt_view_traj}
\end{figure*}

\section{Experiments}
\label{sec:experiments}

\subsection{Experimental setup}
\label{sec:experimental_setup}
\textbf{Datasets:} Evaluation is conducted on two challenging benchmarks: 
\begin{itemize}
\item \textit{Quality-enhanced Oxford RobotCar} ~\cite{RobotCarDatasetIJRR}: Urban driving dataset with 10 km routes under varying weather conditions, processed with ground truth refinement from SGLoc. We use trajectories 11-14-02-26, 14-12-05-52, 14-14-48-55 and 18-15-20-12 as the training set, and trajectories 15-13-06-37, 17-13-26-39, 17-14-03-00 and 18-14-14-42 as the test set.
\item \textit{NCLT} ~\cite{ncarlevaris-2015a}: Campus dataset spanning 5.5 km across seasonal changes, notable for intentional LiDAR vibrations simulating vehicular motion. We use trajectories 2012-01-22, 2012-02-02, 2012-02-18 and 2012-05-11 as the training set, and trajectories 2012-02-12, 2012-02-19, 2012-03-31 and 2012-05-26 as the test set.
\end{itemize}
\textbf{Implementation:} LEADER is implemented in PyTorch ~\cite{Ansel_PyTorch_2_Faster_2024} with MinkowskiEngine ~\cite{choy20194d, choy2019fully, choy2020high, gwak2020gsdn} for sparse convolutions. %Training configurations include:
% \begin{itemize}
% \item Hardware: Intel i9-14900K CPU, NVIDIA RTX 3090 (single) GPU, 128 GB RAM 
% \item Optimization: Adam ~\cite{kingma2017adammethodstochasticoptimization} with initial LR=0.001, multiplicative decay ($\gamma=0.9$)
% \item Geometry: Voxel size $\delta=0.2\:m$, circumferential resolution $L_x=1024$
% \item Training: 50 epochs
% \end{itemize}
Training configurations include: hardware using an Intel i9-14900K CPU, NVIDIA RTX 3090 (single) GPU, and 128 GB RAM; optimization via Adam ~\cite{KingmaB14} with initial learning rate LR=0.001 and multiplicative decay ($\gamma=0.9$); geometric parameters of voxel size $\delta=0.2\:m$ and circumferential resolution $\text{L}_x=1024$; and training duration of 50 epochs.

\subsection{Results on Oxford dataset}
\cref{tab:oxford_results} summarizes the quantitative comparisons on the Quality-enhanced Oxford dataset. LEADER achieves state-of-the-art performance, with the lowest mean position error of 0.63 m and a mean orientation error of 1.11\textdegree, ranking 2nd overall among compared methods and 1st among SCR-based methods. The mean position error represents a 66.1\% reduction over the APR baseline DiffLoc ~\cite{li2024diffloc} and a 24.1\% improvement over the SCR baseline LightLoc ~\cite{li2025lightloc}.

As shown in \cref{fig:oxford_view_traj}, LEADER maintains robust relocalization performance. Our method achieves superior positional precision with minimal catastrophic failures (isolated deviations from ground truth), whereas competitors exhibit more frequent relocalization failures.

\begin{figure}[t]
    %\vspace{-30pt}
    % \small
    \centering
    {\color{gray}
        \fbox{
            \includegraphics[width=0.65\linewidth]{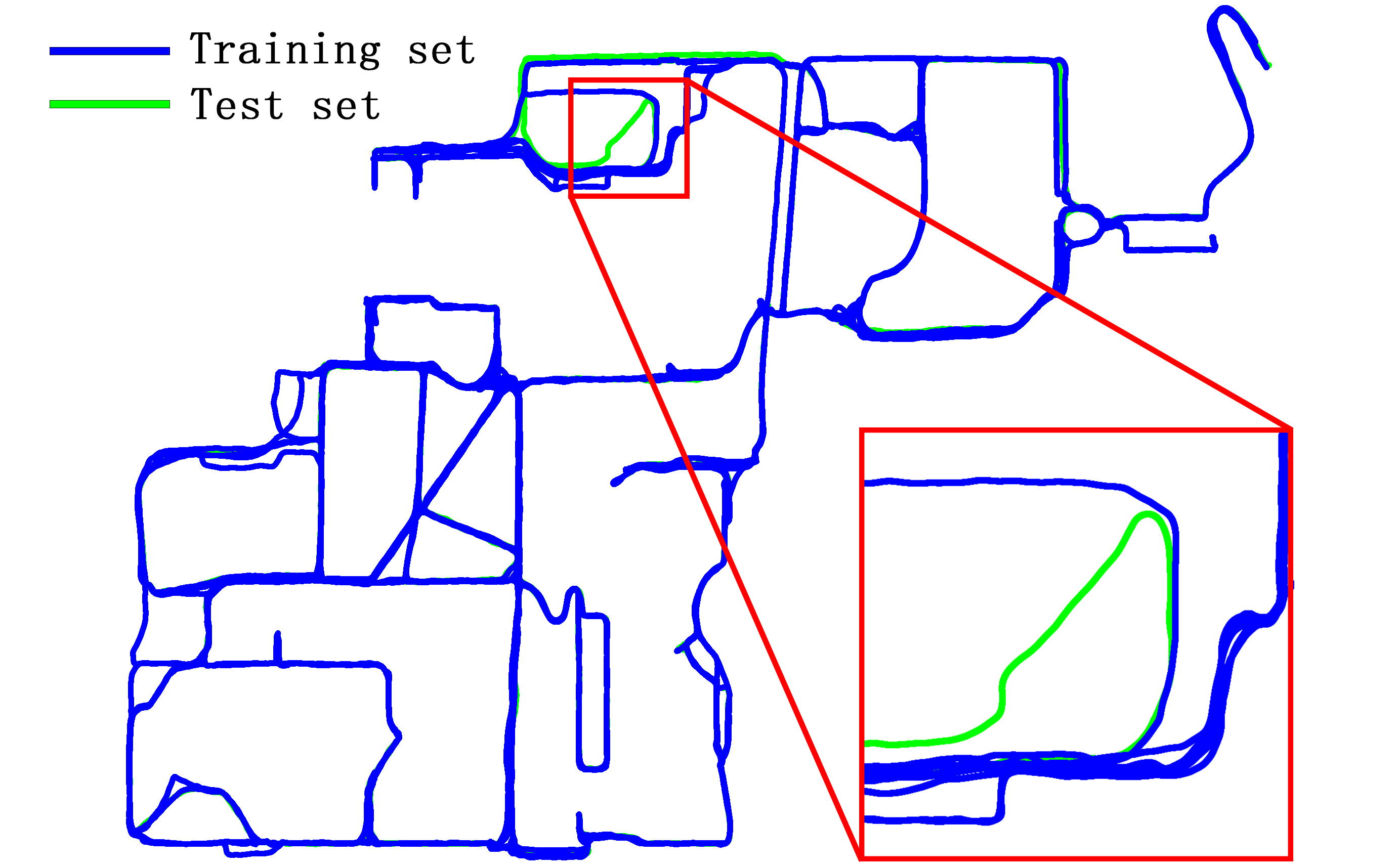}
        }
    }
    \caption{All training trajectories and the 2012-05-26 test trajectory on NCLT dataset.}
    \label{fig:nclt_traj_gt_0526}
\end{figure}

\subsection{Results on NCLT dataset}
\label{sec:nclt}

\cref{tab:nclt_results} demonstrates LEADER's state-of-the-art performance on the high-precision NCLT benchmark, achieving 0.31 m mean position error and 1.81\textdegree mean orientation error. This represents a substantial improvement over existing methods, reducing position error by 73.9\% compared to the APR baseline DiffLoc and by 79.5\% against the SCR baseline LiSA. To our knowledge, LEADER is the first implicit LiDAR-based relocalization method achieving sub-0.5 m precision on NCLT dataset, overcoming challenges of sparse vegetation and seasonal variations.

\cref{fig:nclt_view_traj} visualizes results from challenging trajectory 2012-05-26, where training and test trajectories exhibit significant non-overlapping regions (\cref{fig:nclt_traj_gt_0526}). Despite significant viewpoint mismatches, our method maintains 0.32 m positional accuracy, surpassing DiffLoc and LiSA by 83.0\% and 90.3\% respectively. This robustness stems from two key factors: 
\begin{itemize}
\item \textbf{Enhanced geometric resilience:} Spatial Transformation maintains stable representations across diverse angle.
\item \textbf{Reliability filtering:} The TRR loss effectively suppresses unreliable points, ensuring only high-reliability features contribute to pose estimation.  
\end{itemize}

\begin{figure}
    \small
    \centering
    \includegraphics[width=1\linewidth]{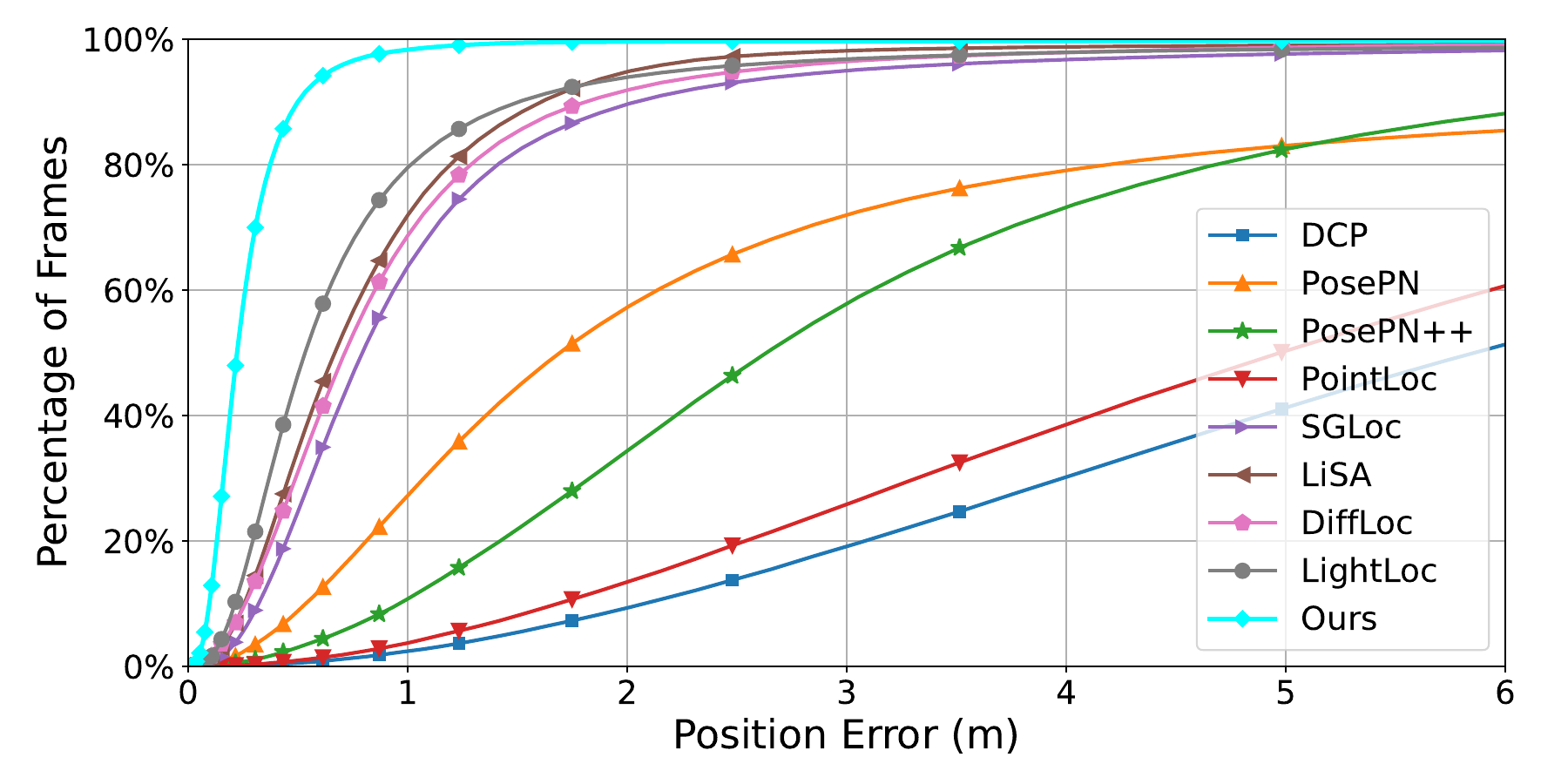}
    \caption{Cumulative distribution of position errors on NCLT dataset.}
    \label{fig:frame_loss_cmp}
\end{figure}

\begin{table}[t]
    \centering
    \small
    \caption{Frame coverage comparison at critical error thresholds.}
    \begin{tabular}{lccc}
    \toprule
    Method & \multicolumn{1}{c}{\textless{}0.5m (\%)} & \multicolumn{1}{c}{\textless{}1m (\%)} & \multicolumn{1}{c}{99\% Thresholds (m)} \\
    \midrule
    SGLoc & 24.5 & 63.8 & 8.70 \\
    DiffLoc & 30.8 & 68.7 & 5.73 \\
    LiSA & 34.3 & 72.0 & 4.98 \\
    LightLoc & 46.3 & 79.6 & 8.70 \\
    \textbf{Ours} & \textbf{90.0} & \textbf{98.3} & \textbf{1.23} \\
    \bottomrule
    \end{tabular}
    \label{tab:coverage_comparison}
\end{table}

\begin{table*}[ht]
  \centering
  \small
  % \setlength{\tabcolsep}{1mm}
  % \vspace{0.01\textwidth}
  \caption{Ablation study evaluating the contributions of the RPGE and TRR modules on NCLT dataset. The ST column indicates the use of Spatial Transformation, while TRR-Train and TRR-Test denote the application of the  TRR module during training and inference, respectively. Results for both normal and yaw-perturbed point clouds are reported as position/orientation errors (m, \textdegree), with inference time (ms) and parameters (M is million) included as an additional metric.}
  \begin{tabular}{lccccccc}
    \toprule
    Method & ST & TRR-Train & TRR-Test & Normal Point Cloud & Yaw-Perturbed Point Cloud & Inference Time & Parameters \\
    \midrule
    1 & \texttimes & \texttimes & \texttimes & 0.95, 1.94 & 8.04, 10.77 & 48 & 69.67 M \\
    2 & \texttimes & \checkmark & \texttimes & 0.89, 1.82 & 8.67, 10.44 & 48 & 69.67 M \\
    3 & \texttimes & \checkmark & \checkmark & 0.88, 1.86 & 6.59, 10.14 & 48 & 69.67 M \\
    4 & \checkmark & \texttimes & \texttimes & 0.59, 2.12 & 0.59, 2.12 & 48 & 69.67 M \\
    5 & \checkmark & \checkmark & \texttimes & 0.35, 1.80 & 0.35, 1.80 & 48 & 69.67 M \\ 
    6 & \checkmark & \checkmark & \checkmark & 0.31, 1.81 & 0.31, 1.81 & 48 & 69.67 M \\
    \bottomrule
  \end{tabular}
  \label{tab:ablation}
\end{table*}

To further demonstrate relocalization consistency, \cref{fig:frame_loss_cmp} presents the cumulative distribution of position errors across all test frames. LEADER achieves exceptional coverage: 90.0\% of frames localized within 0.5m error and 98.3\% within 1m. 

As shown in \cref{tab:coverage_comparison}, this significantly outperforms existing methods at critical precision thresholds. Notably, competitors require 4.98-8.70m error thresholds to reach 99\% frame coverage, while LEADER achieves this milestone at just 1.23m. This unprecedented consistency confirms our method's capability to deliver reliable sub-meter precision in challenging environments.

\subsection{Runtime}
\label{sec:runtime}
LEADER achieves real-time performance on both the Oxford RobotCar (20 Hz) and NCLT (10 Hz) datasets. It processes each frame in 46 ms and 48 ms respectively, which is below the acquisition interval of each dataset. This ensures efficient real-time operation suitable for autonomous navigation.

%% file: sec/5_ablation.tex
\subsection{Ablation studies}
\label{sec:ablation}

Based on \cref{tab:ablation}, we analyze the contributions of the RPGE and TRR modules on NCLT dataset:

\textbf{Robust Projection-based Geometric Encoder (RPGE)}:
The Spatial Transformation component enhances robustness against yaw variations. Without this component (Methods 1–3), performance significantly degrades under perturbations (\eg, 8.04 m vs. 0.95 m position error for Method 1). In contrast, when enabled (Methods 4–6), the framework maintains consistent accuracy across perturbation conditions (\eg, 0.59 m, 2.12\textdegree for both cases in Method 4). This demonstrates the component's effectiveness in stabilizing representations against environmental variations.

\textbf{Truncated Relative Reliability (TRR)}:
TRR introduces reliability-aware weighting during training (TRR-Train) and reliability-based pose refinement during inference (TRR-Test). Without TRR (Methods 1 and 4), baselines achieve 0.95 m (Method 1) and 0.59 m (Method 4) position errors. Enabling TRR-Train alone (Methods 2 and 5) reduces errors to 0.89 m and 0.35 m, respectively, demonstrating adaptive weighting improves feature learning. Further enabling TRR-Test (Methods 3 and 6) yields additional gains, as reliability-based refinement suppresses outlier predictions. We observe limited orientation error improvement, potentially due to sparser point distributions in distant regions after Spatial Transformation (~\cref{fig:project_comparation}), as orientation estimation relies more on long-range features.

\textbf{Efficiency}:
A critical outcome of our ablation study is the negligible runtime cost and fewer than 0.001 million addition parameters associated with the RPGE and TRR modules. Despite the inference time remaining unchanged, the relocalization accuracy is profoundly enhanced.

\begin{figure}[t!]
    \centering
    \includegraphics[width=\linewidth]{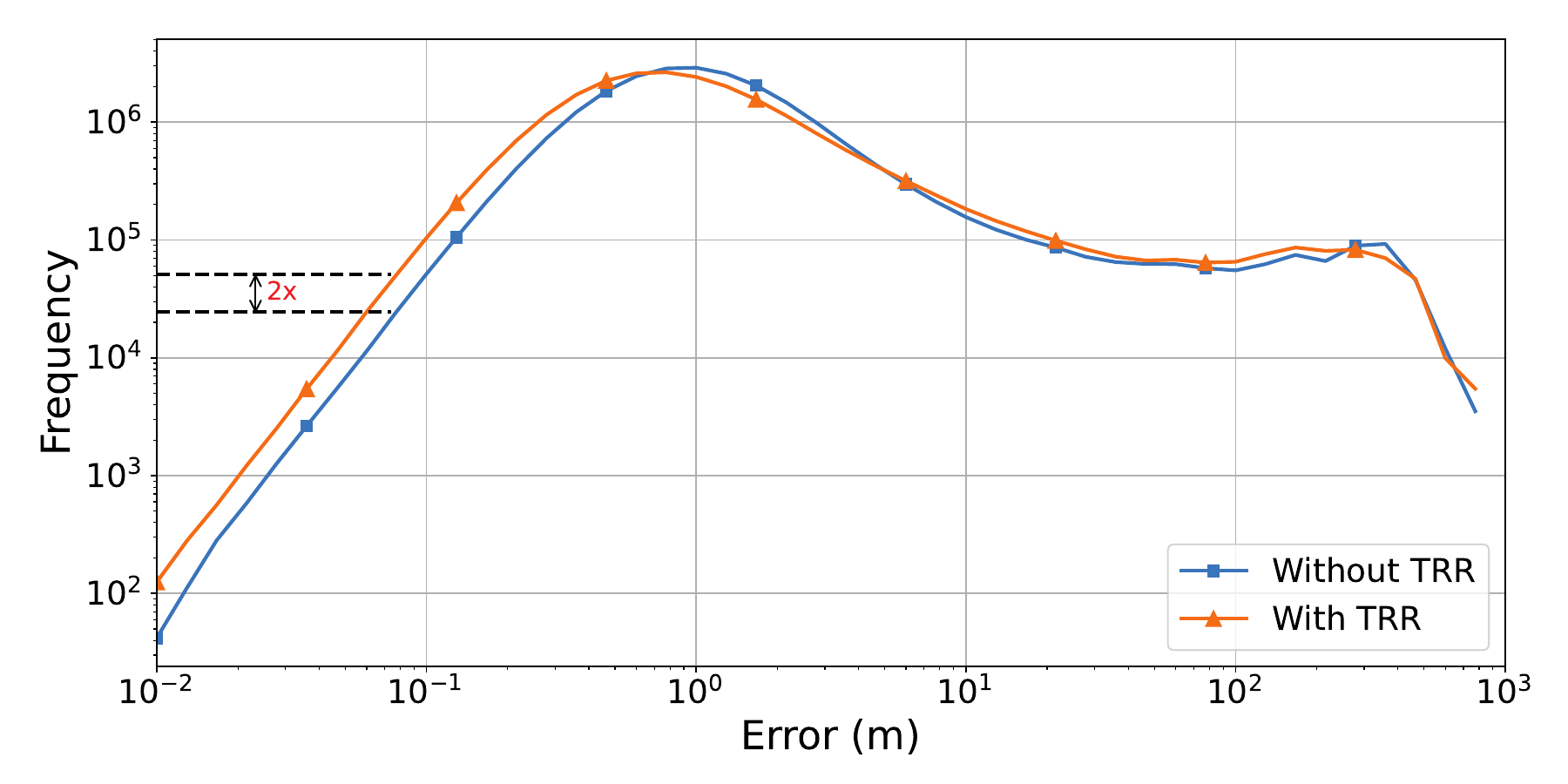}
    \caption{TRR Loss effect on Scene Point Error distribution on NCLT dataset 2012-05-26 trajectory.}
    \label{fig:side:trc_error_distri}
\end{figure}

The supplementary results in \cref{fig:side:trc_error_distri} and \cref{fig:side:cfd_error_distri} further validate the mechanisms of the TRR module:

\begin{figure}[t!] 
    \centering
    \includegraphics[width=\linewidth]{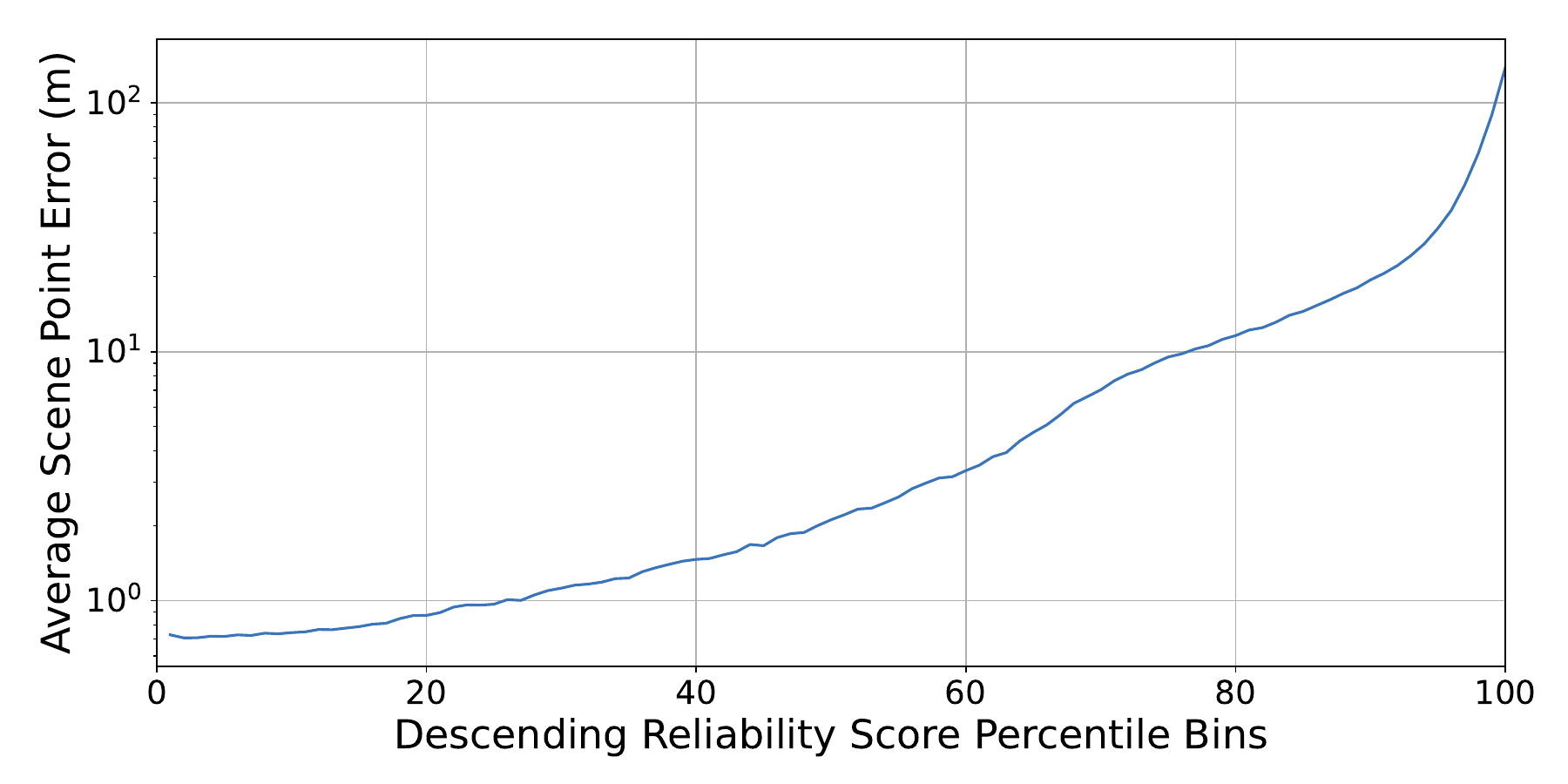}
    \caption{Average Scene Point Error vs. Reliability Score Percentiles on NCLT dataset 2012-05-26 trajectory.}
    \label{fig:side:cfd_error_distri}
\end{figure}

\textbf{Error distribution shift (\cref{fig:side:trc_error_distri})}:
Enabling TRR significantly reshapes the scene point error distribution. The proportion of high-precision points (\textless{} 0.3 m) \textbf{doubles} compared to the baseline without TRR, show TRR prioritizes accurate relocalization for reliable regions. Meanwhile, the increase in \textgreater{} 5 m (outliers) suggests TRR implicitly downweights challenging areas during training, aligning with its reliability-aware loss mechanism. These outliers are suppressed during inference by RANSAC, which discards low-reliability predictions. This trade-off reflects the fixed model capacity: TRR allocates resources to optimize learnable regions while tolerating irreducible errors in ambiguous areas.

\textbf{Reliability-error correlation (\cref{fig:side:cfd_error_distri})}:
The monotonic relationship between reliability scores and scene point errors confirms TRR’s ability to estimate prediction reliability. Points with top reliability scores (lowest percentile) achieve an average error of 0.4~m, whereas the lowest-reliability points exhibit catastrophic errors (\textgreater{}100 m). This justifies our inference strategy of retaining only the top-k reliable points for pose estimation, as low-reliability regions introduce noise that degrades overall accuracy.

\textbf{Robustness analysis (\cref{tab:robust})}:
We further evaluate the proposed method under several degraded sensing scenarios to assess robustness. The baseline uses standard training and testing on NCLT, while other rows introduce test-time perturbations. Under extreme occlusion simulated by a 180° frontal field-of-view (FOV), the system maintains reasonable pose estimation despite a noticeable performance drop. When subjected to random point dropout of up to 50\%, only minor degradation is observed, indicating strong resilience to point cloud sparsity. The addition of Gaussian noise ($\sigma = 0.05$) applied to all points in the point cloud results in a slight performance degradation, with the error closely matching the baseline, indicating negligible impact under low-noise conditions. To evaluate the system’s inherent tolerance to orientation variations without relying on ground point correction, which is typically used to level the point cloud, we introduce pitch and roll perturbations of up to $10^\circ$. The results demonstrate that the method remains reliable under such perturbations, which simulate typical vehicle vibrations. Notably, in all perturbed scenarios, our approach achieves relocalization accuracy superior to the previous state-of-the-art result of (1.19 m, 2.31°) obtained under normal testing conditions, underscoring the robustness and practical applicability of the proposed system.

\begin{table}[t]
    \small
    \centering
    \setlength{\tabcolsep}{6.5mm}
    \caption{Robustness evaluation under various degraded sensing conditions on the NCLT dataset. All values represent the mean position and orientation errors.}
    \begin{tabular}{lc}
        \toprule
         Test Condition & Error (m, \textdegree) \\
         \midrule
         Baseline & 0.31, 1.81 \\
         180\textdegree\,FOV & 0.78, 3.51 \\
         Random Dropout (0 - 50\%) & 0.33, 1.86 \\
         Gaussian Noise ($\sigma=0.05$) & 0.36, 1.93 \\
         Pitch/Roll Perturbation ($\pm10^{\circ}$) & 0.65, 3.17 \\
         \bottomrule
    \end{tabular}
    \label{tab:robust}
\end{table}

%% file: sec/6_conclusion.tex
\section{Conclusion}
\label{sec:conclusion}

In this paper, we propose LEADER, a robust LiDAR-based global relocalization framework that significantly advances scene coordinate regression performance. Our approach introduces two key innovations: 1) A Robust Projection-based Geometric Encoder establishing robust scene representations resilient to yaw variations, and 2) A Truncated Relative Reliability loss enabling reliable estimation. Extensive experiments demonstrate state-of-the-art performance across challenging datasets, with LEADER achieving unprecedented 0.31~m positional accuracy on the NCLT benchmark - the first implicit relocalization method to attain sub-0.5~m precision on this dataset.

Unlike prior methods that require full trajectory and rotation alignment between training and test data, our framework only requires trajectory proximity, greatly enhancing applicability. Currently, our method focuses on handling yaw angle variations, with pitch and roll effects mitigated through ground-based correction. Future work will aim to address the full SE(3) relocalization problem without relying on ground plane detection.

\noindent\textbf{Acknowledgements.} This work was supported by the National Natural Science Foundation of China (No.~62501502). 

%% file: sec/X_suppl.tex
\clearpage
\setcounter{page}{1}
\maketitlesupplementary

\begin{figure}[t]
\centering
\includegraphics[width=\linewidth]{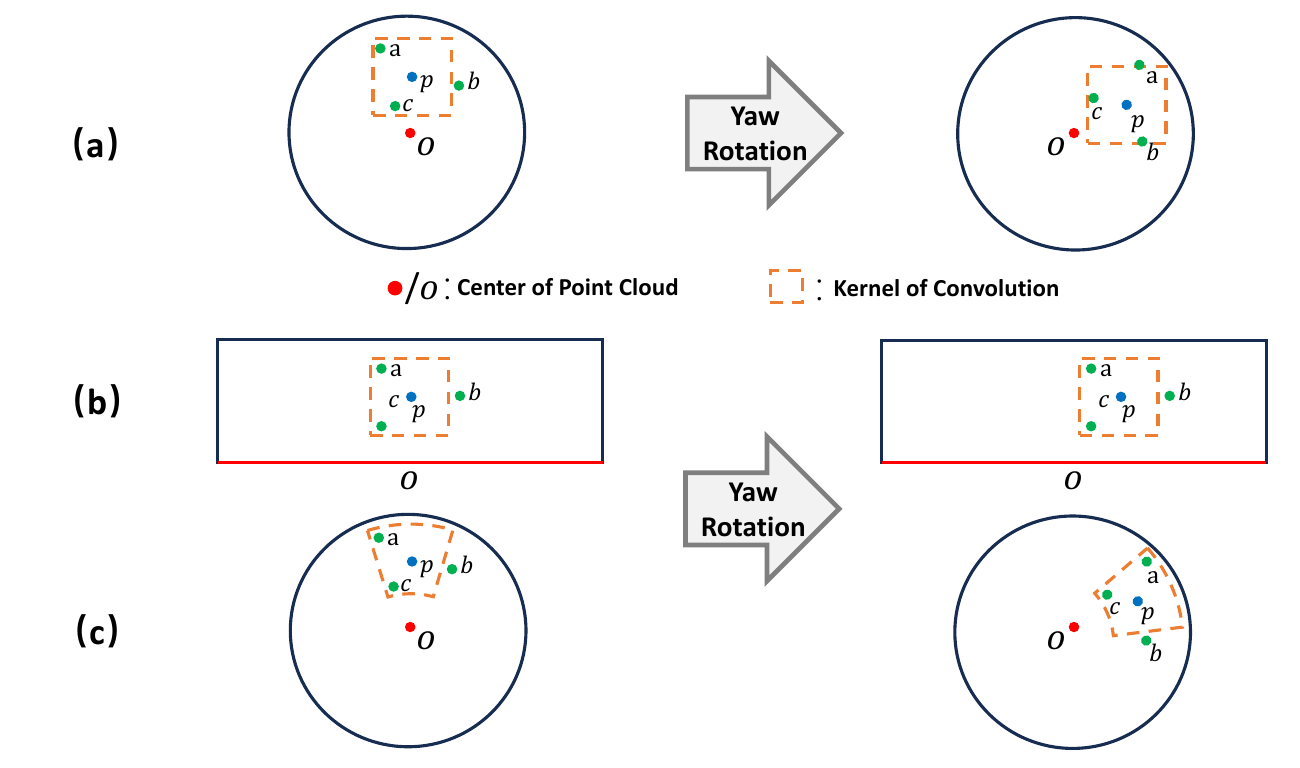}
\caption{
Illustration of cylindrical projection and convolution correspondence.
\textbf{(a)} Original point cloud $\mathcal{P}_1$ in Cartesian coordinates.
\textbf{(b)} Projected point cloud $\mathcal{P}_2$ in cylindrical coordinates after Spatial Transformation.
\textbf{(c)} The equivalent convolution operation in $\mathcal{P}_1$ corresponding to the rectangular convolution kernel applied on $\mathcal{P}_2$, which forms a sector-shaped receptive field.
}
\label{fig:projection_illustration}
\end{figure}

\section{Supplement for Method}
\subsection{The planar rectification in Spatial Transformation}

Given the raw point cloud $\mathcal{P}_{\text{raw}} = \{\mathbf{p}_i = (x_i, y_i, z_i)\}_{i=1}^N$ and the corresponding transformation matrix $\mathbf{T}$, we first perform the ground plane rectification using Patchwork++ \cite{lee2022patchworkpp}:

\begin{enumerate}
\item Estimate ground plane equation $ax + by + cz + d = 0$ via RANSAC\cite{Zhou2018}:
\item Compute rotation matrix $\mathbf{R}_{\text{plane}}$ aligning ground normal $\mathbf{n} = (a,b,c)$ with z-axis:
\begin{equation}
\begin{aligned}
\theta &= \arccos\left(\frac{\mathbf{n} \cdot \mathbf{e}_z}{\|\mathbf{n}\|}\right), \quad
% \end{aligned}
% \end{equation}
% \begin{equation}
% \begin{aligned}
\mathbf{v} = \frac{\mathbf{n} \times \mathbf{e}_z}{\|\mathbf{n} \times \mathbf{e}_z\|}, \\
% \end{aligned}
% \end{equation}
% \begin{equation}
% \begin{aligned}
\mathbf{R}_{\text{plane}} &= \exp(\theta \mathbf{v}_\times) \quad \text{(Rodrigues' formula)}.
\end{aligned}
\label{eq:rot_plane}
\end{equation}

\item We denote the transformation matrix for the planar rectification as $\mathbf{T}_{\text{plane}}$:
\begin{equation}
\mathbf{T}_{\text{plane}} = \begin{bmatrix}
\mathbf{R}_{\text{plane}} & \mathbf{t}_{\text{plane}} \\
\mathbf{0} & \mathbf{1}
\end{bmatrix}, \quad
% \end{equation}
% \begin{equation}
\mathbf{t}_{\text{plane}} = \frac{-d}{\|\mathbf{n}\|^2}\mathbf{n}.
\label{eq:trans_plane}
\end{equation}

\item We apply the following rectification to correct the point cloud to horizontal: 
\begin{equation}
\mathbf{p}_i' = \mathbf{R}_{\text{plane}} \cdot \mathbf{p}_i + \mathbf{t}_{\text{plane}}.
\label{eq:rectify}
\end{equation}
\end{enumerate}

\begin{figure}[t]
    \centering
    \includegraphics[width=0.95\linewidth]{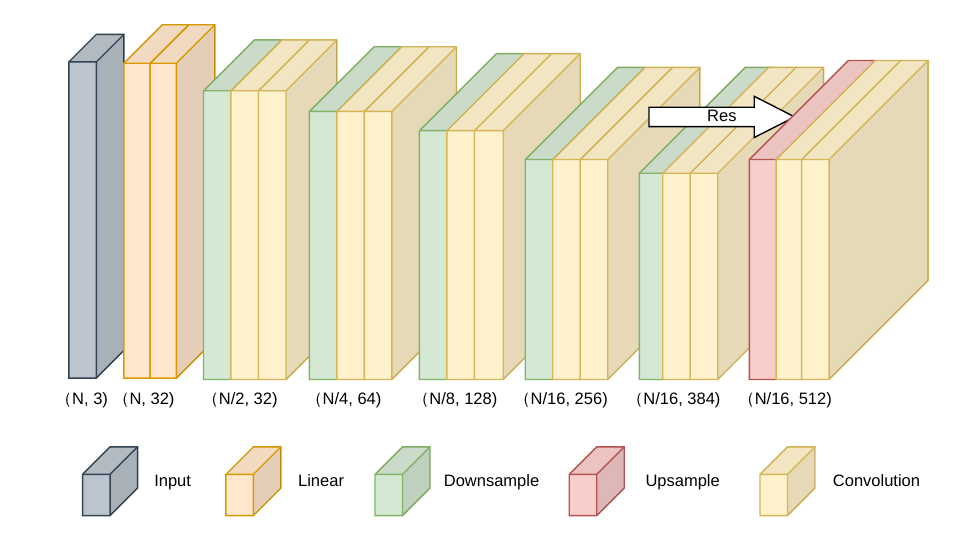}
    \caption{Overview of the proposed encoder architecture.}
    \label{fig:detail_encoder}
\end{figure}

\begin{figure}[htbp]
    \centering
    \includegraphics[width=0.95\linewidth]{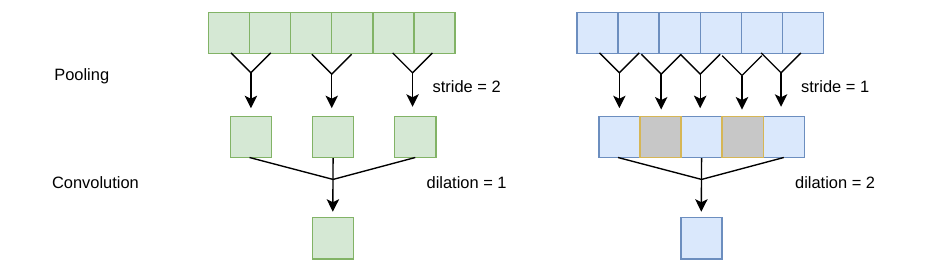}
    \caption{Illustration of the dilated cyclic sparse convolution used in stage 5 to enlarge the receptive field without reducing spatial resolution (right).}
    \label{fig:dilation_convolution}
\end{figure}

\subsection{Detailed Encoder Architecture}

As illustrated in Figure~\ref{fig:detail_encoder}, our encoder begins with a 3-dimensional input, which is first projected to 16 dimensions using a residual fully connected layer, and then further projected to 32 dimensions. The architecture then consists of six stages: stages 1–5 perform downsampling followed by two convolutional layers, and stage 6 performs upsampling, fusion, and two convolutional layers.

For stages 1–4, downsampling is implemented with a cyclic sparse convolution using stride=2, kernel=2, and dilation=1, whose output is concatenated with the result of max pooling. This is followed by two cyclic sparse convolution layers with stride=1, kernel=3, and dilation=1. After the fourth downsampling stage, the point cloud has been downsampled by a factor of 16, resulting in relatively sparse points. In stage 5, to further enlarge the receptive field without reducing the number of points, we set stride=1 and increase dilation to 2 in the cyclic sparse convolution, as illustrated in Figure~\ref{fig:dilation_convolution}.

Stage 6 performs upsampling using a transposed cyclic sparse convolution with stride=1, kernel=2, and dilation=1. The upsampled features are then concatenated with the output from stage 4 and projected to 512 dimensions. Finally, two cyclic sparse convolution layers with stride=1, kernel=3, and dilation=2 are applied to produce the final enriched features.

\subsection{Regressor hyperparameters}
The design of our regression network is inspired by the FFN layers in the Transformer architecture and the commonly used pooling layers in the field of computer vision, and the hyperparameters, including the number of heads and layers, are set based on empirical values.

\begin{table}[t!]
  \centering
  \small
  \caption{Localization results for the NCLT dataset at different yaw angles. All values are given as (m, °).}
  % \setlength{\tabcolsep}{7pt}
  % \vspace{0.015\textwidth}
  % \setlength{\tabcolsep}{3pt}
  \begin{tabular}{lcccc}
    \toprule
    Yaw Angle & LEADER & LightLoc & DiffLoc \\
    \midrule
    Original & 0.31, 1.81 & 1.70, 2.95 & 1.19, 2.31 \\
    Yaw +180\textdegree & 0.31, 1.81 & 21.70, 21.57 & 5.58, 36.36 \\
    Random Yaw & 0.31, 1.81 & 30.51, 22.46 & 4.57, 76.50 \\
    \bottomrule
  \end{tabular}
  \label{tab:st_test}
\end{table}
  
\section{RPGE Module Analysis}
\subsection{Projection Description for Yaw Robustness}
In the Spatial Transformation module, we convert the Cartesian coordinates to the Cylindrical coordinates. In \cref{eq:geo_proj}, the $z$-axis remains unchanged during projection; thus, we omit the $z$-axis (equivalent to a top-down view) and voxelization to illustrate the principle of this projection. The visualization is shown in \cref{fig:projection_illustration}. Let the original point cloud be $\mathcal{P}_1$ and the projected point cloud be $\mathcal{P}_2$. The kernel of sparse convolution is rectangular. When applying convolution on the projected point cloud $\mathcal{P}_2$, the rectangular kernel corresponds to a sector in the original point cloud $\mathcal{P}_1$. When a yaw rotation occurs in $\mathcal{P}_1$, the point cloud rotates clockwise or counterclockwise. In contrast, in $\mathcal{P}_2$, this rotation is transformed into a translation. Convolution inherently possesses translation equivariance. Consider the changes before and after rotation at a point $p$. In $\mathcal{P}_1$, both the region covered by the convolutional kernel at point $p$ and the corresponding points at different kernel positions change, leading to a lack of rotation robustness. However, in $\mathcal{P}_2$, the region and relative relationships covered by the convolutional kernel at point $p$ remain consistent before and after rotation, thereby endowing the method with yaw robustness.

Furthermore, after projection, the originally continuous yaw angles in $\mathcal{P}_1$ become discontinuous at the two ends of $\mathcal{P}_2$. To address this, we employ circular sparse convolution by pre-padding both sides according to the kernel size. After convolution, the padded regions are removed, as \cref{eq:cyclic_padding}. This ensures full-range yaw robustness.

\subsection{Robustness Analysis of Spatial Transformation}

To validate the effectiveness of the Spatial Transformation module, we conduct comprehensive comparisons including our method and two state-of-the-art methods (LightLoc~\cite{li2025lightloc} and DiffLoc~\cite{li2024diffloc}) under varying yaw rotations on the NCLT dataset~\cite{ncarlevaris-2015a}. As shown in Table~\ref{tab:st_test}, the proposed Spatial Transformation demonstrates remarkable robustness to yaw variations:

\begin{itemize}
    \item \textbf{Compared methods} exhibit comparable sensitivity to yaw angle variations: LightLoc achieves a performance of 21.70~m / 21.57° at yaw+180° and 30.51~m / 22.46° under random yaw, whereas DiffLoc attains 5.58~m / 36.36° and 4.57~m / 76.50° under the same respective conditions.
    
    \item \textbf{Our method} maintains consistent performance across all yaw angles, achieving stable errors of 0.31~m/1.81° regardless of the rotation magnitude.
\end{itemize}

These results conclusively demonstrate that our Spatial Transformation module is essential for achieving yaw-invariant localization, effectively mitigating performance degradation under rotational variations commonly encountered in real-world autonomous driving scenarios.
\begin{table}[t!]
    \centering
    \small
    \setlength{\tabcolsep}{1.5mm}
    \caption{Performance comparison of coordinate systems}
    % \vspace{-10pt}
    % \vspace{0.01\textwidth}
    \begin{tabular}{lcc}
        \toprule
        Method & Position Error & Orientation Error\\
        \midrule
        Cylindrical projection & 0.31~m & 1.81\textdegree\\
        Spherical projection & 0.35~m (\textuparrow~12.9\%) & 1.97\textdegree(\textuparrow~8.8\%) \\
        \bottomrule
    \end{tabular}
    \label{tab:coord_ablation}
\end{table}

\subsection{Coordinate System Ablation Study}  
Table~\ref{tab:coord_ablation} compares our Sptial Transformation (Cylindrical projection) with Spherical projection:

The Spherical Projection exhibits higher errors than our Spatial Transformation approach, with a +12.9\% increase in position error (0.35 vs. 0.31~m) and a +8.8\% increase in orientation error (1.97° vs. 1.81°). 

\section{TRR Module Analysis}
\subsection{TRR Loss Principle}
We hereby elucidate the principle of our proposed Truncated Relative Reliability (TRR) loss. The Euclidean loss serves as the fundamental point-wise loss, as defined in \cref{eq:rawloss} of the main text.

We perform two normalization operations on the reliability scores, given by \cref{eq:scale,eq:trunc} in the main text, where the constant $K_s = \frac{\ln 10}{\pi}$.

The core components of TRR loss are defined in \cref{eq:point_weight,eq:finaloss} of the main text.

In \cref{eq:scale,eq:trunc}, we employ $\arctan$ to constrain values within $[-\frac{\pi}{2}, \frac{\pi}{2}]$, and scale them by $K_s$. The rationale for choosing $K_s = \frac{\ln 10}{\pi}$ is that it expands the output range of $u_i$ to $[-\frac{\ln 10}{2}, \frac{\ln 10}{2}]$. After exponentiation in \cref{eq:point_weight}, this range becomes $[10^{-\frac{1}{2}}, 10^{\frac{1}{2}}]$, where the maximum value is 10 times the minimum. Therefore, the constant $K_s$ constrains the ratio between maximum and minimum weights in \cref{eq:point_weight} to approximately 10. This prevents the weights from degenerating into the mean Euclidean loss when differences are too small, while also avoiding scenarios where the model focuses excessively on high-quality points, neglecting moderately effective points during training.

Furthermore, we employ two components for reliability normalization: $u_i^{\text{scale}}$ and $u_i^{\text{cut}}$. In \cref{eq:point_weight}, $w_i$ is normalized using both $u_i^{\text{scale}}$ and $u_i^{\text{cut}}$.

When $u_i \in [-10\pi, 10\pi]$, $u_i^{\text{scale}} = u_i^{\text{cut}}$, and the weight calculation $w_i$ becomes equivalent to softmax. Considering the total loss formulation $\mathcal{L}_{\text{TRR}} = \sum_i w_i \mathcal{L}_{\text{raw},i}$, we analyze the contribution of a specific point $m$. The total loss can be decomposed as:
\begin{equation}
    \mathcal{L}_{\text{TRR}} = w_m \cdot \mathcal{L}_{\text{raw},m} + \sum_{j \neq m} w_j \cdot \mathcal{L}_{\text{raw},j}
\end{equation}
where $\sum_{j \neq m} w_j = 1 - w_m$ due to the softmax-like normalization. For analytical clarity, we consider the average behavior of other points by defining $\bar{\mathcal{L}}_{\text{raw}}^{m-} = \frac{\sum_{j \neq m} w_j \mathcal{L}_{\text{raw},j}}{\sum_{j \neq m} w_j}$, allowing us to express the loss as:
\begin{equation}
    \mathcal{L}_{\text{TRR}} \approx w_m \cdot \mathcal{L}_{\text{raw},m} + (1 - w_m) \cdot \bar{\mathcal{L}}_{\text{raw}}^{m-}
\end{equation}
This approximation highlights the trade-off between point $m$ and other points. When point $m$ has high feature quality and consequently better regression quality ($\mathcal{L}_{\text{raw},m} < \bar{\mathcal{L}}_{\text{raw}}^{m-}$), the model can reduce the total loss by increasing $u_m$, which increases $w_m$ and decreases the relative contribution of other points. Conversely, when point $m$ has low quality, decreasing $u_m$ reduces the total loss. This compels the model to prioritize learning high-quality points, while the $K_s$ constant ensures all points receive adequate training by constraining the weight range.

When $u_i$ exceeds $[-10\pi, 10\pi]$, we analyze the case where $u_i > 10\pi$. Consider point $n$ with weight $w_n$ and Euclidean loss $\mathcal{L}_{\text{raw},n}$, while other points have losses $\mathcal{L}_{\text{raw},i}$ and weights $w_i$. The total loss becomes:
\begin{equation}
    \mathcal{L}_{\text{TRR}} = w_n \cdot \mathcal{L}_{\text{raw},n} + \sum_{i \neq n} w_i \cdot \mathcal{L}_{\text{raw},i}
\end{equation}
When $u_n > 10\pi$, $u_n^{\text{scale}}$ continues to increase while $u_n^{\text{cut}}$ remains constant, resulting in $u_n^{\text{scale}} > u_n^{\text{cut}}$. In \cref{eq:point_weight}, the numerator increases while the denominator remains unchanged, causing $w_n$ to increase. Since the denominator remains fixed, other weights $w_i$ stay constant, leading to $\sum w_i > 1$ and consequently an increased total loss. To minimize loss, the model avoids $u$ exceeding $10\pi$. The case for $u_n < -10\pi$ follows similar reasoning.

The dual design of $u_i^{\text{scale}}$ and $u_i^{\text{cut}}$ ensures gradients exist even when $u_i$ exceeds $[-10\pi, 10\pi]$. This avoids the gradient vanishing problem that would occur with only $u_i^{\text{scale}}$ due to floating-point precision limitations, while also preventing the complete absence of gradients that would result from using only $u_i^{\text{cut}}$ beyond the clamping range.

\subsection{Reliability Visualization}
\begin{figure}[t]
    \centering
    {\color{gray}\fbox{
        \includegraphics[width=0.85\linewidth]{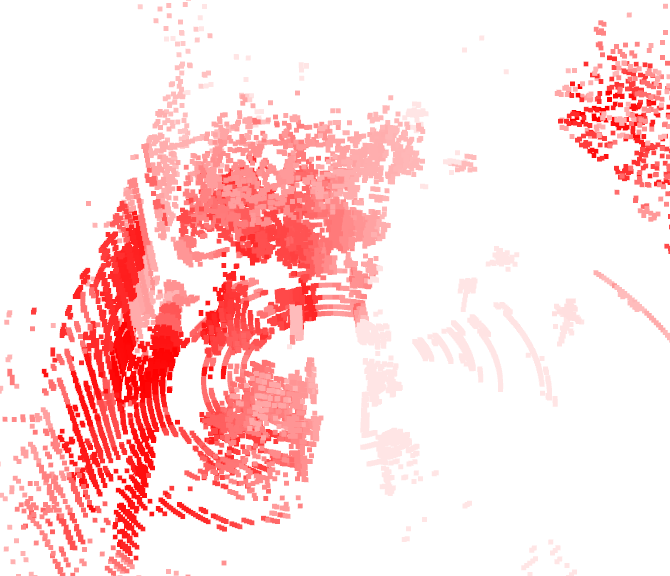}
    }}
    \caption{Visualization of Reliability scores. Points are color-coded by reliability (dark red: high, light/white: low).}
    \label{fig:reliability_vis}
\end{figure}

We visualize the reliability scores $u$, as shown in Fig.~\ref{fig:reliability_vis}. The point cloud is color-coded by reliability, where darker red indicates higher reliability and lighter/white colors indicate lower reliability. 

As observed, points on buildings and at the base of dense vegetation exhibit higher reliability, while ground points, sparse shrubs, and vegetation tops (leaves) show lower reliability. This demonstrates the model's ability to effectively distinguish between different structural elements in the environment. 

Notably, we observe that ground points adjacent to buildings or dense vegetation also maintain relatively high reliability. This can be attributed to the receptive field of sparse convolution - the stacked convolutional layers in the encoder ensure that each point represents features from its local neighborhood. Thus, even ground points incorporate contextual information from nearby structures, enhancing their reliability.

\subsection{Quantitative Analysis by Reliability Groups}
To further validate the effectiveness of reliability scores, we partition points into quartiles based on their reliability rankings and evaluate pose estimation performance using each subset independently. The results are presented in Table~\ref{tab:reliability_groups}.

\begin{table}[t]
    \centering
    % \small
    \caption{Mean position error (m) and mean orientation error (°) across reliability quartiles.}
    \setlength{\tabcolsep}{6.5mm}
    \label{tab:reliability_groups}
    \begin{tabular}{lcc}
        \toprule
        Reliability Quartile & Error (m/°) \\
        \midrule
        0-25\% (Highest) & 0.32, 1.96 \\
        25-50\% & 0.34, 1.97 \\
        50-75\% & 0.67, 2.33 \\
        75-100\% (Lowest) & 2.39, 3.63 \\
        \bottomrule
    \end{tabular}
\end{table}

The results clearly demonstrate that higher reliability correlates with better pose estimation accuracy. The highest reliability group achieves position and orientation errors of 0.32~m and 1.96° and respectively, while the lowest reliability group shows significantly higher errors of 2.39~m and 3.63°. This represents a 86.7\% reduction in position error and a 46.0\% reduction in orientation error for the highest reliability group compared to the lowest reliability group.

\begin{table}[t]
    \centering
    \small
    \caption{Performance improvement by integrating TRR loss into SGLoc}
    % \vspace{-10pt}
    % \vspace{0.01\textwidth}
    \setlength{\tabcolsep}{3pt}
    \begin{tabular}{lcc}
        \toprule
        Metric & Position Error (m) & Orientation Error (°) \\
        \midrule
        SGLoc (Original) & 1.83 & 3.54 \\ 
        SGLoc + TRR & 1.76 (\textdownarrow~3.83\%) & 2.61 (\textdownarrow~26.27\%) \\
        \bottomrule
    \end{tabular}
    \label{tab:trr_ablation_sgloc}
\end{table}

\subsection{TRR Loss Generalizability Analysis}  
To demonstrate the portability of our proposed TRR loss, we integrate it into SGLoc~\cite{li2023sgloc}. As shown in Table~\ref{tab:trr_ablation_sgloc}, quantitative improvements are observed as follows:

\begin{itemize}
    \item \textbf{Position error:} Reduced from 1.83~m to 1.76 m (3.83\% reduction)
    \item \textbf{Orientation error:} Reduced from 3.54\textdegree~to 2.61\textdegree\,(26.27\% reduction)
\end{itemize}

These results confirm that TRR loss is not merely specialized for our architecture, but serves as a generalizable strategy for other localization architecture.

\subsection{Comparison with Alternative Loss Functions}

To validate the effectiveness of our proposed TRR loss, we compare it against two alternative loss functions: the standard Euclidean distance mean loss and a modified version of the Matching loss from RSKDD-Net. Given ground truth coordinates $\mathbf{c}_i^{\text{gt}}$, predicted coordinates $\mathbf{\hat{c}}_i$, and reliability scores $u_i$, we evaluate the following loss functions:

\begin{itemize}
    \item \textbf{Mean Euclidean loss}: This baseline loss function is defined as the mean of the per-point losses $\mathcal{L}_{\text{raw},i}$ over the entire point cloud:
    \begin{equation}
        \mathcal{L}_{\text{raw}} = \frac{1}{N}\sum_i^N \mathcal{L}_{\text{raw},i}
    \end{equation}
    where N is the total number of points.
    \item \textbf{Modified Matching loss}: We adapt the RSKDD-Net~\cite{Lu_2020_NeurIPS} Matching loss to our framework. The formulation is:
    \begin{equation}
        w_i = \max\{\sigma^{\max} - \sigma_i, 0.01\}, \quad \tilde{w_i} = \frac{w_i}{\sum_j w_j}
    \end{equation}
    \begin{equation}
        \mathcal{L}_{\text{matching}} = \sum_i \tilde{w_i} \|\mathbf{c}_i^{\text{gt}} - \mathbf{\hat{c}}_i\|_2
    \end{equation}
    We set the hyperparameter $\sigma^{\max} = 1$.
    
    \item \textbf{TRR loss}: Our proposed loss as defined in \cref{eq:point_weight} and \cref{eq:finaloss} of the main text.
\end{itemize}

The quantitative comparison on the NCLT dataset is presented in \cref{tab:loss_comparison}. All methods maintain identical inference time (48 ms), ensuring fair comparison.

\begin{table}[t]
    \centering
    \small
    \caption{Performance comparison of different loss functions on NCLT dataset}
    \setlength{\tabcolsep}{10pt}
    \label{tab:loss_comparison}
    \begin{tabular}{lcc}
        \toprule
        Method & Error (m, °) \\
        \midrule
        Mean Euclidean loss & 0.59, 2.12 \\
        Matching loss & 0.43, 1.97 \\
        \textbf{TRR loss (Ours)} & \textbf{0.31, 1.81} \\
        \bottomrule
    \end{tabular}
\end{table}

Our TRR loss demonstrates significant performance improvements over alternative approaches. Compared to the mean Euclidean loss, TRR achieves 47.4\% and 14.6\% reduction in position and orientation errors respectively. When compared to Matching loss, TRR reduces errors by 27.9\% and 8.1\%.

\begin{figure}[t]
    \begin{subfigure}{0.48\linewidth}
        {\color{gray}
            \fbox{\includegraphics[width=0.93\linewidth]{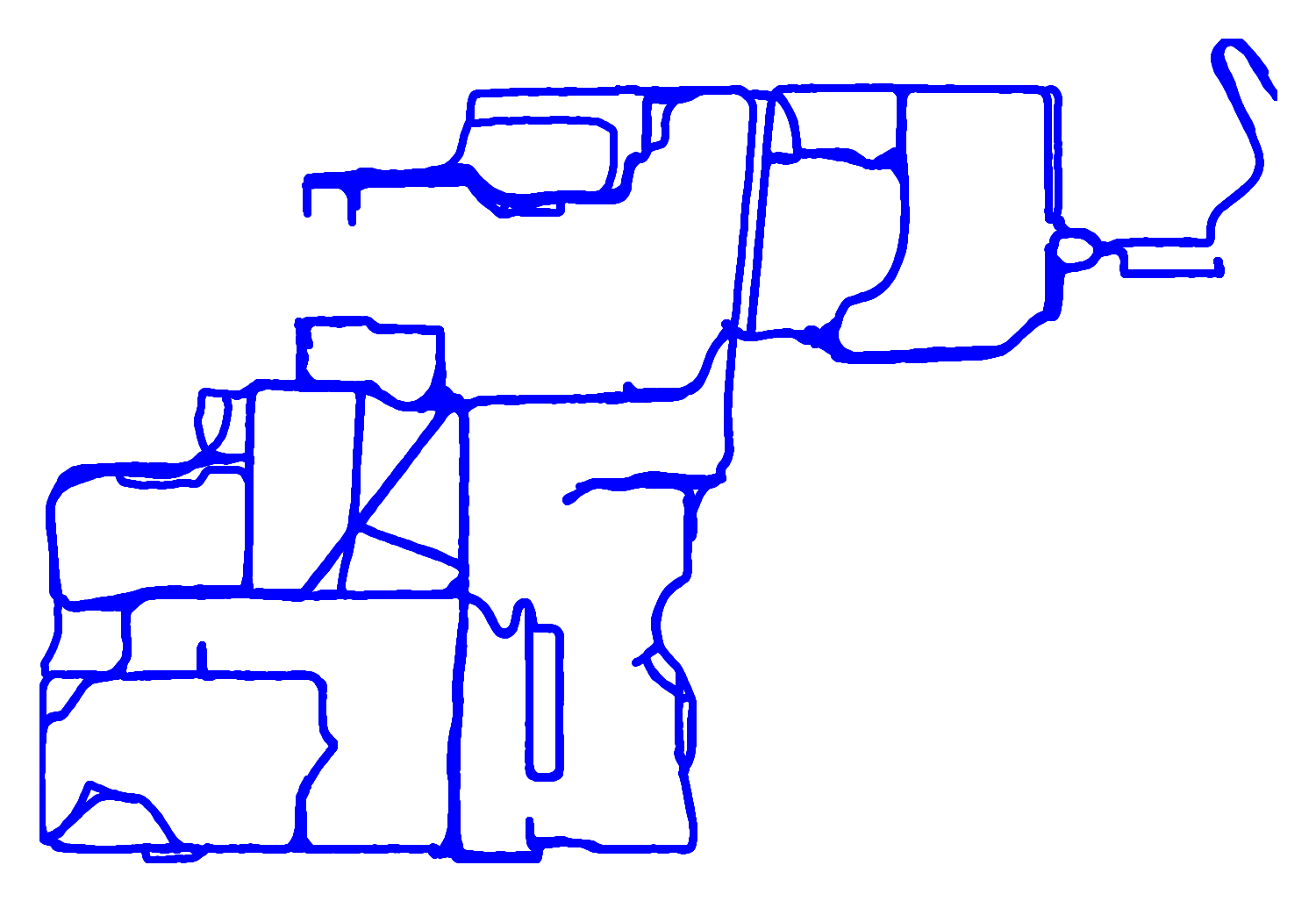}}
        }
        \caption{training set}
        \label{fig:nclt_trainig_set}
    \end{subfigure}
    \hfill
    \begin{subfigure}{0.48\linewidth}
        {\color{gray}
            \fbox{\includegraphics[width=0.93\linewidth]{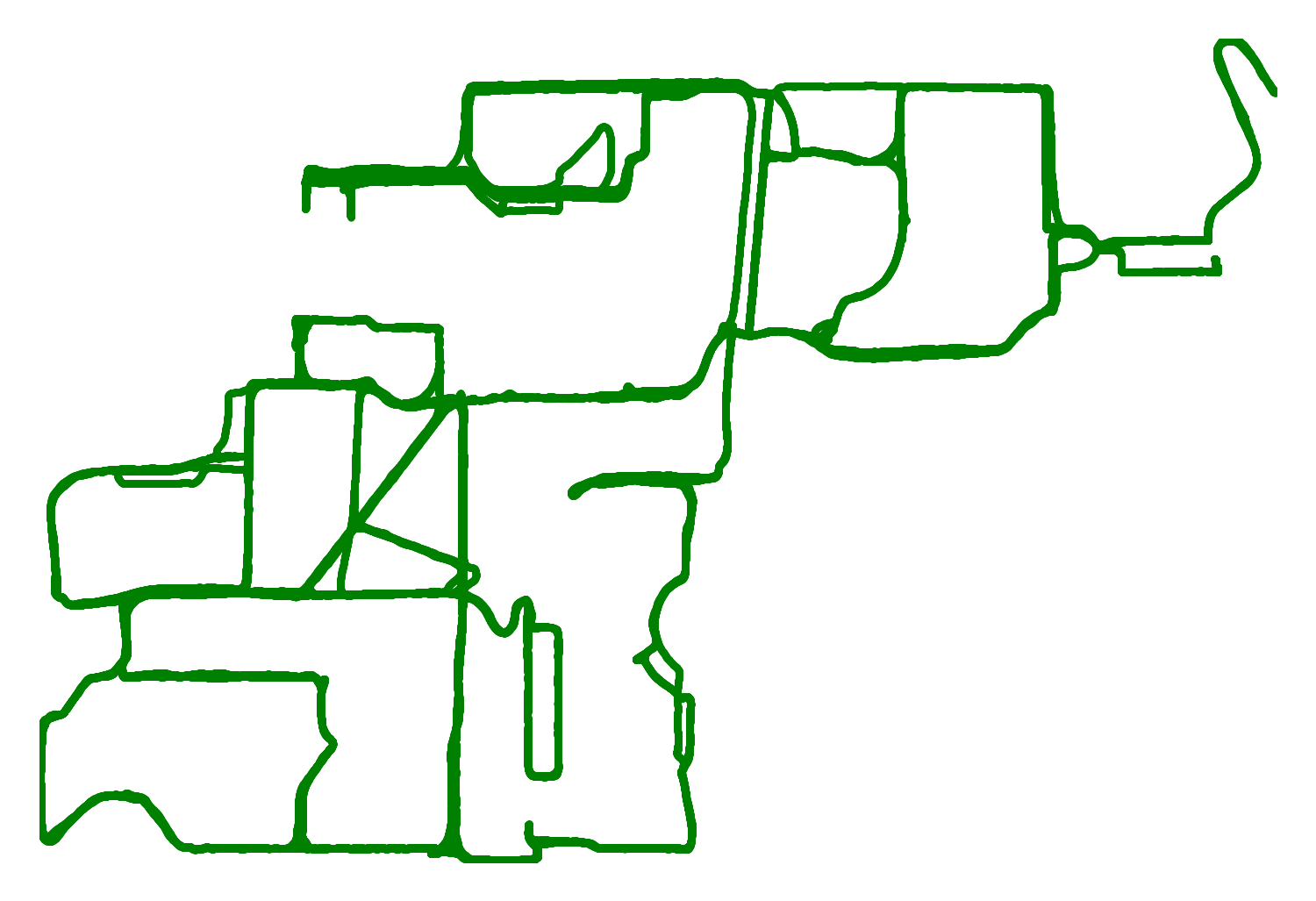}}
        }
        \caption{test set}
        \label{fig:nclt_test_set}
    \end{subfigure}
    \caption{Trajectory visualization of the NCLT dataset.}
\label{fig:dataset_split}
\end{figure}

\section{Evaluation Metrics}
Let $\mathbf{T}_{\text{final},i}$ (\cref{eq:pose_comp}) denote the estimated pose transformation matrix of the $i$-th frame and $\mathbf{T}_{\text{gt},i}$ represent its ground-truth pose transformation matrix. The evaluation metrics are defined as:

\textbf{Mean Position Error:}  
For each frame $i$, let $\mathbf{t}_{\text{final},i} \in \mathbb{R}^3$ and $\mathbf{t}_{\text{gt},i} \in \mathbb{R}^3$ be the translation vectors extracted from $\mathbf{T}_{\text{final},i}$ and $\mathbf{T}_{\text{gt},i}$, respectively. The Mean Position Error is computed as:
\begin{equation}
    \text{MPE} = \frac{1}{N} \sum_{i=1}^{N} \left\| \mathbf{t}_{\text{final},i} - \mathbf{t}_{\text{gt},i} \right\|_2
\end{equation}
where $N$ is the total number of frames.

\textbf{Mean Orientation Error:}  
For each frame $i$, let $\mathbf{R}_{\text{final},i}$ and $\mathbf{R}_{\text{gt},i}$ be the rotation matrices from $\mathbf{T}_{\text{final},i}$ and $\mathbf{T}_{\text{gt},i}$, respectively. The Mean Orientation Error error (in degrees) is calculated using:
\begin{equation}
    \text{MOE} = \frac{1}{N} \sum_{i=1}^{N} \left( \frac{180}{\pi} \cdot \arccos \left( \frac{\text{tr}(\mathbf{R}_{\text{rel},i}) - 1}{2} \right) \right)
\end{equation}
where $\mathbf{R}_{\text{rel},i} = (\mathbf{R}_{\text{final},i})^\top \mathbf{R}_{\text{gt},i}$ and $\text{tr}(\cdot)$ denotes the matrix trace. The error is bounded to $[0^\circ, 180^\circ]$.

\begin{table}[t]
\centering
\small
\setlength{\tabcolsep}{5pt}
\caption{Comparisons of different methods on the NCLT dataset.}
\label{tab:retrieval_compare}
\begin{tabular}{@{}lcccc@{}}
\toprule
\textbf{Metric} & \textbf{LEADER} & \textbf{LEADER (xy)} & \textbf{RING/RING++} \\
\midrule
MPE (m) & 0.31 & 0.28 & 16.34 \\
MOE (°) & 1.81 & 1.03 & 10.45 \\
MedPE (m) & 0.24 & 0.21 & 0.27 \\
Recall@1 & — & — & 75.07\% \\
Success@5m & 99.72\% & 99.72\% & 92.79\% \\
\bottomrule
\end{tabular}
\end{table}
\section{Additional experiment}
We compare our method against the retrieval-based approaches RING ~\cite{lu2022one} and RING++ ~\cite{ring++} on the NCLT dataset. The same mapping trajectories (with keyframes selected at 10 m intervals) and test trajectories as used in the main paper are adopted for these baselines, while the training and test trajectories for our method follow the same configuration as described in the Experiment section of the main paper. \textbf{All frames in the test set are used for evaluation}. 

To enable a direct comparison, we additionally report the performance of our method in the xy-plane. Specifically, we compute the position error as the Euclidean distance error in xy coordinates, and the orientation error as the yaw angle difference. This is because the baseline methods directly provide localization results in the xy-plane. Table~\ref{tab:retrieval_compare} summarizes the quantitative results. In the table, MPE, MOE, and MedPE denote Mean Position Error, Mean Orientation Error, and Median Position Error, respectively. For retrieval methods, Recall@1 indicates retrieval success within 5 m, while Success@5m represents final localization success after registration. For our method, Success@5m directly measures localization accuracy within 5m. For each metric, we report the best result achieved by either RING or RING++.

Based on the results in Table~\ref{tab:retrieval_compare}, our method achieves an MPE of 0.28 m and an MOE of 1.03° in the xy-plane, substantially outperforming RING/RING++, which yield 16.34 m and 10.45°, respectively. We attribute this large margin in part to the high failure rate of the retrieval-based baselines, which significantly inflates their average errors. Specifically, RING/RING++ attain a Recall@1 of only 75.07\%, and after registration, their Success@5m reaches 92.79\%, still leaving a failure rate of 7.21\%. In contrast, our method achieves a Success@5m of 99.72\%, with a failure rate of only 0.28\%, which is less than 3.9\% of that of the RING/RING++. Moreover, in terms of MedPE, which is largely unaffected by failed frames, our method achieves 0.21 m, still clearly outperforming the 0.27 m of RING/RING++.

\begin{figure}[t]
    \centering
    {\color{gray}
        \fbox{\includegraphics[width=0.8\linewidth]{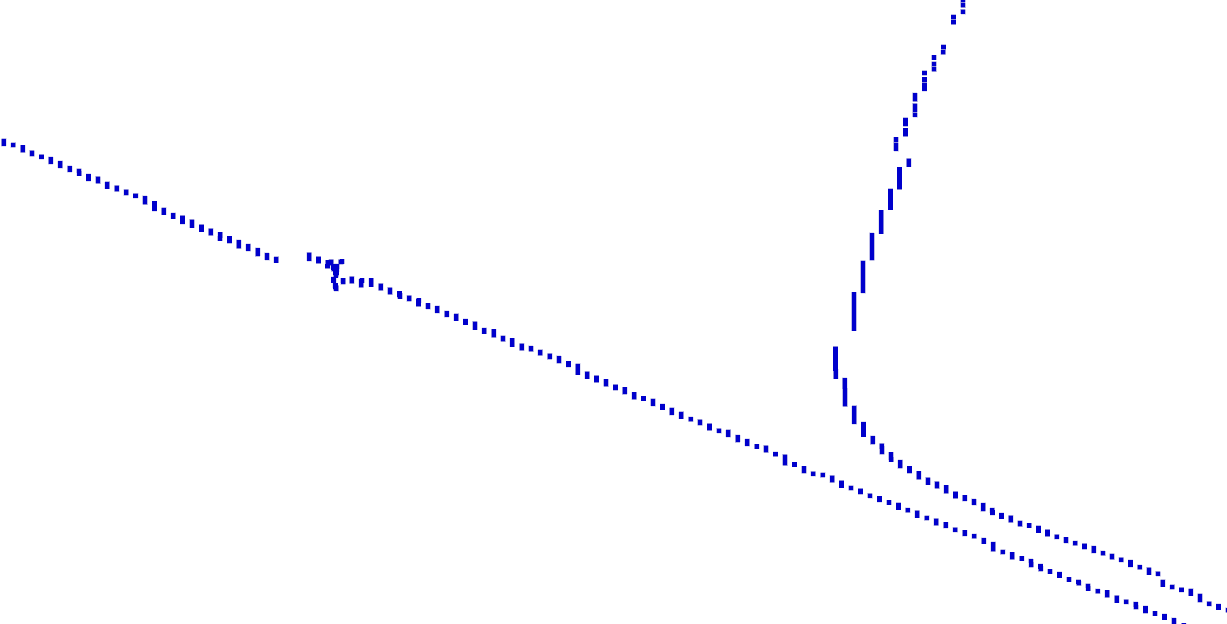}}}
    \caption{Ground truth trajectory visualization of Quality-enhanced Oxford dataset.}
    \label{fig:oxford_local_gt}
\end{figure}

\section{Dataset}
\subsection{Dataset Selection}
We justify our dataset selection by addressing the absence of KITTI~\cite{Geiger2012CVPR}, a common LiDAR SLAM benchmark. Our task involves relocalization within a pre-built map, which requires significant overlap between training and test trajectories in \textbf{seen} scenes. However, most KITTI sequences lack such overlap, making it suboptimal for this purpose. Consequently, we use the NCLT and Oxford datasets, where training and test data are collected in the same environment with substantial trajectory overlap. Fig.~\ref{fig:dataset_split} visualizes this overlap in the NCLT dataset, demonstrating the high spatial coincidence essential for our evaluation.

\subsection{Analysis on Quality-enhanced Oxford Dataset}
While our method achieves a 74.9\% reduction in mean position error on the NCLT dataset compared to state-of-the-art methods, the improvement on the Quality-enhanced Oxford dataset~\cite{RobotCarDatasetIJRR} is 24.9\%. Although still significantly superior, this performance gap warrants further analysis.

We attribute this discrepancy primarily to limitations in the ground truth of the Quality-enhanced Oxford dataset. Despite the calibration improvements introduced by SGLoc, the corrected trajectory exhibits noticeable jagged artifacts and discontinuities due to its floating-point precision limitations, as visualized in Fig.~\ref{fig:oxford_local_gt}. These irregularities in the ground truth trajectory inevitably constrain further accuracy improvements, as they introduce inherent uncertainties during both training and test phases.

\section{Visualization}
A more comprehensive comparison between Quality-enhanced Oxford and NCLT is presented in \cref{fig:vis_nclt,fig:vis_oxford}. As can be seen, LEADER has almost no catastrophic errors in its estimates, proving the robustness of its localization.
% \newpage

\begin{table*}
\centering
\small
\setlength{\tabcolsep}{15pt}
\begin{tabularx}{\textwidth}{rccccc}
    & 15-13-06-37 & 17-13-26-39 & 17-14-03-00 & 18-14-14-42 \\
    
    \raisebox{6ex}{\rotatebox[]{90}{PNVLAD}} &
    \includegraphics[width=0.154\linewidth]{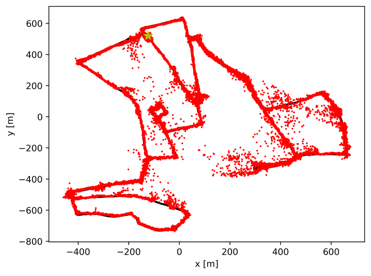} &
    \includegraphics[width=0.154\linewidth]{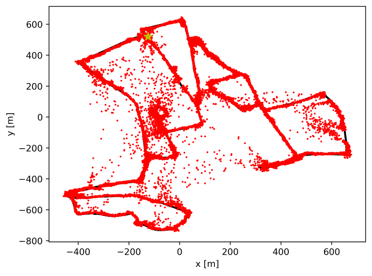} &
    \includegraphics[width=0.154\linewidth]{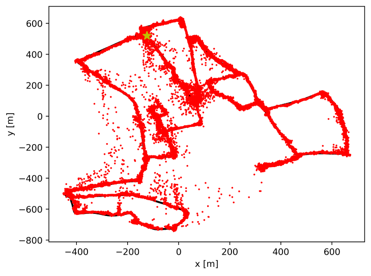} &
    \includegraphics[width=0.154\linewidth]{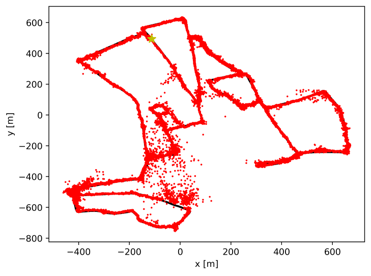} \\
    
    \raisebox{6ex}{\rotatebox[]{90}{DCP}} &
    \includegraphics[width=0.154\linewidth]{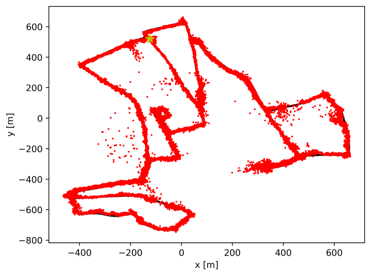} &
    \includegraphics[width=0.154\linewidth]{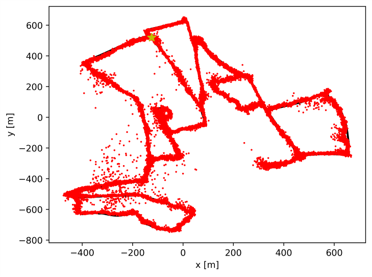} &
    \includegraphics[width=0.154\linewidth]{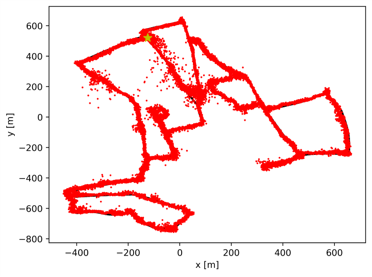} &
    \includegraphics[width=0.154\linewidth]{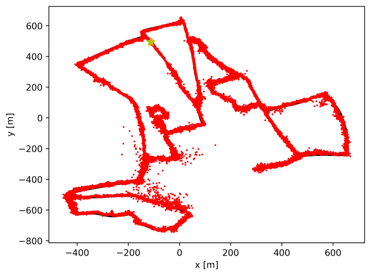} \\
    
    % \raisebox{6ex}{\rotatebox[]{90}{PosePN}} &
    % \includegraphics[width=0.154\linewidth]{sec/diagram/qeoxford0637_posepn.png} &
    % \includegraphics[width=0.154\linewidth]{sec/diagram/qeoxford2639_posepn.png} &
    % \includegraphics[width=0.154\linewidth]{sec/diagram/qeoxford0300_posepn.png} &
    % \includegraphics[width=0.154\linewidth]{sec/diagram/qeoxford1442_posepn.png} \\
    
    \raisebox{6ex}{\rotatebox[]{90}{PosePN++}} &
    \includegraphics[width=0.154\linewidth]{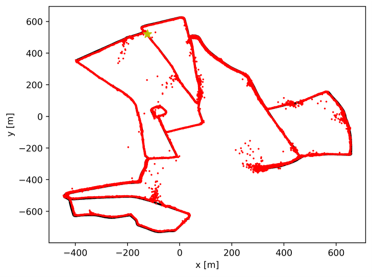} &
    \includegraphics[width=0.154\linewidth]{sec/diagram/qeoxford2639_posepnpp.png} &
    \includegraphics[width=0.154\linewidth]{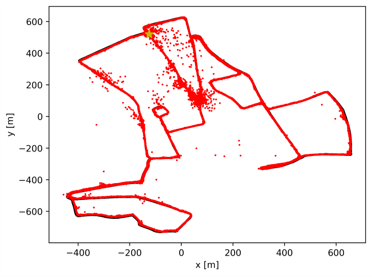} &
    \includegraphics[width=0.154\linewidth]{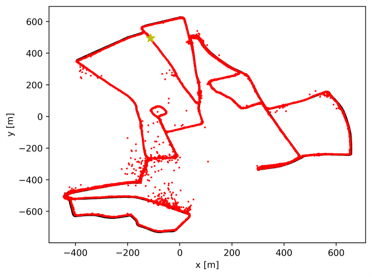} \\
    
    \raisebox{6ex}{\rotatebox[]{90}{PointLoc}} &
    \includegraphics[width=0.154\linewidth]{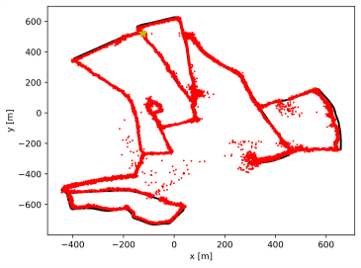} &
    \includegraphics[width=0.154\linewidth]{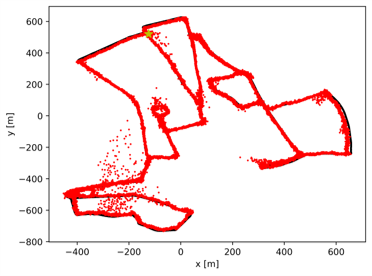} &
    \includegraphics[width=0.154\linewidth]{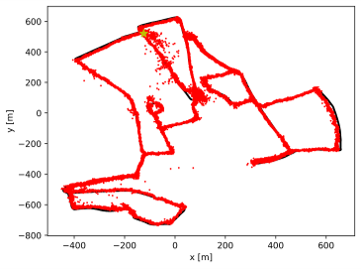} &
    \includegraphics[width=0.154\linewidth]{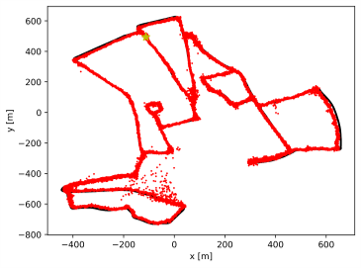} \\
    
    \raisebox{6ex}{\rotatebox[]{90}{HypLiLoc}} &
    \includegraphics[width=0.154\linewidth]{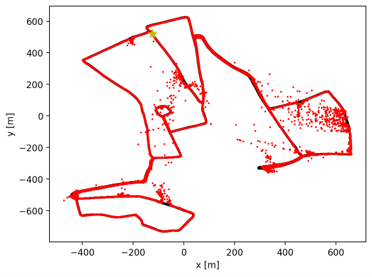} &
    \includegraphics[width=0.154\linewidth]{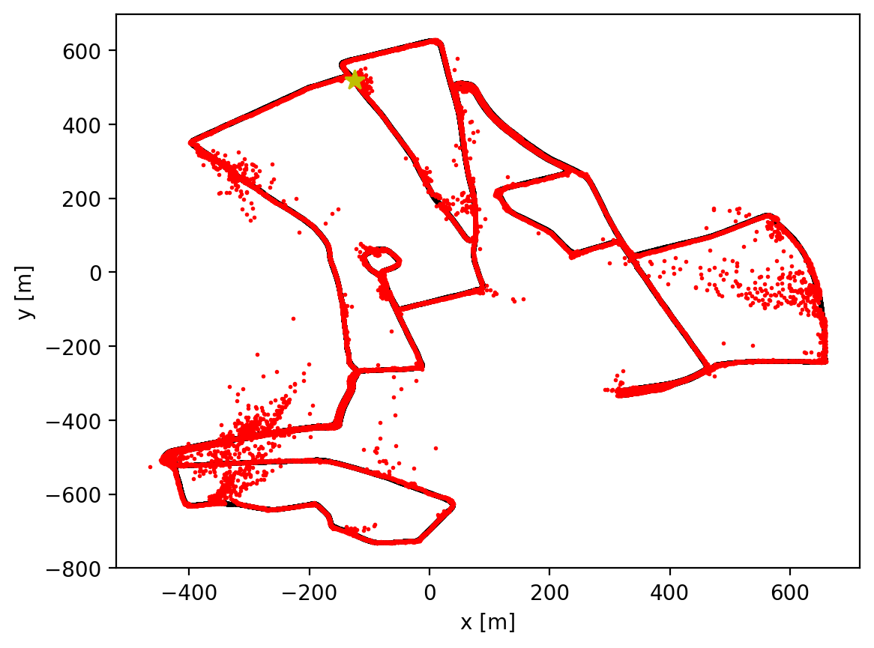} &
    \includegraphics[width=0.154\linewidth]{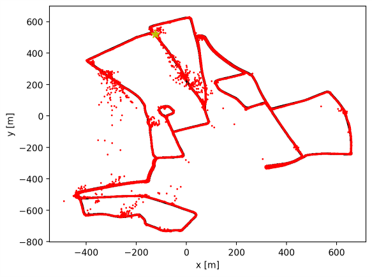} &
    \includegraphics[width=0.154\linewidth]{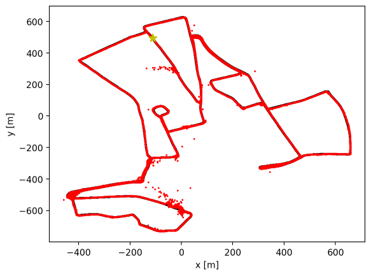} \\
    
    \raisebox{6ex}{\rotatebox[]{90}{SGLoc}} &
    \includegraphics[width=0.154\linewidth]{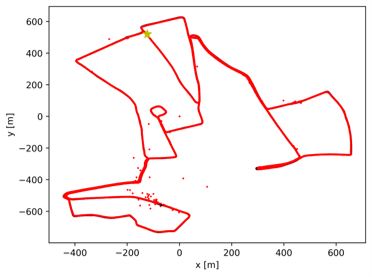} &
    \includegraphics[width=0.154\linewidth]{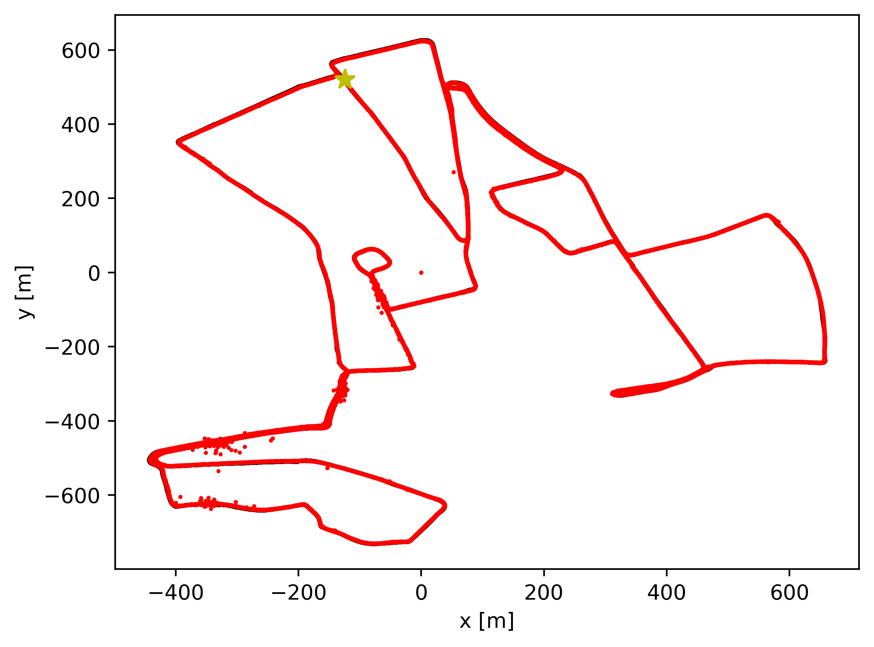} &
    \includegraphics[width=0.154\linewidth]{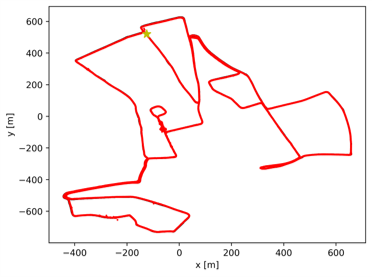} &
    \includegraphics[width=0.154\linewidth]{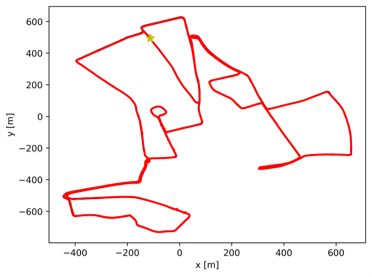} \\
    
    \raisebox{6ex}{\rotatebox[]{90}{LiSA}} &
    \includegraphics[width=0.154\linewidth]{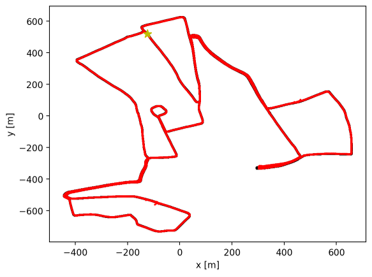} &
    \includegraphics[width=0.154\linewidth]{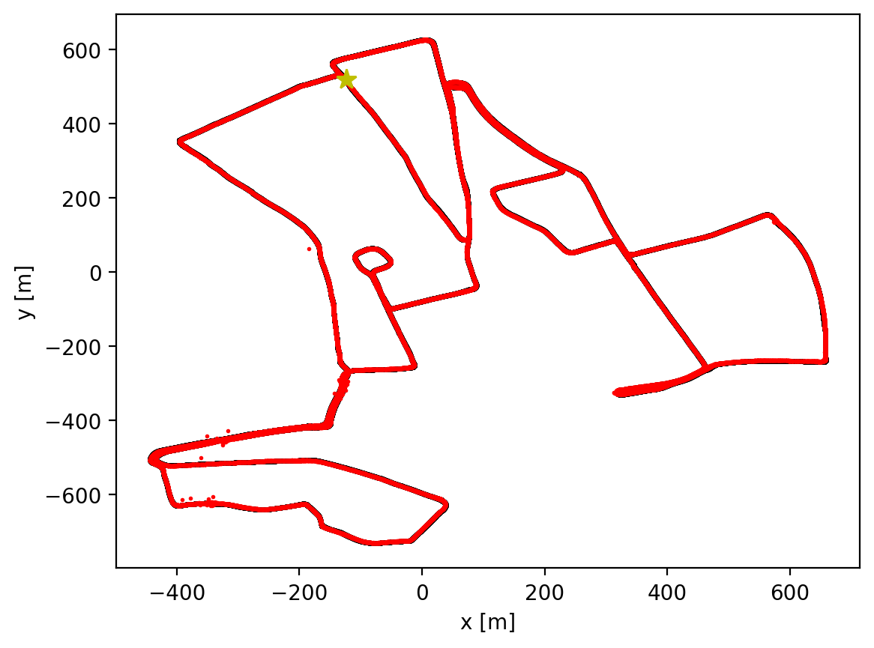} &
    \includegraphics[width=0.154\linewidth]{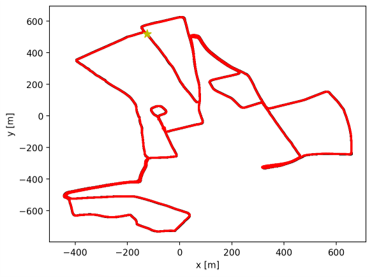} &
    \includegraphics[width=0.154\linewidth]{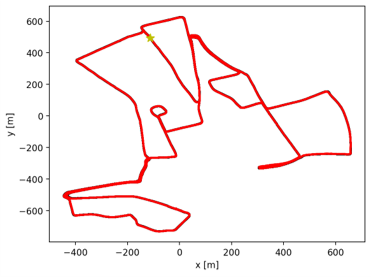} \\
    
    \raisebox{6ex}{\rotatebox[]{90}{DiffLoc}} &
    \includegraphics[width=0.154\linewidth]{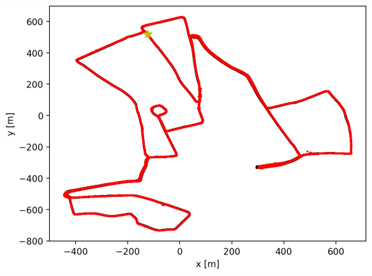} &
    \includegraphics[width=0.154\linewidth]{sec/diagram/qeoxford2639_diffloc.png} &
    \includegraphics[width=0.154\linewidth]{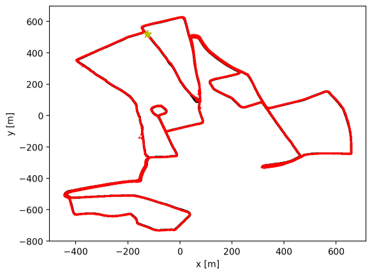} &
    \includegraphics[width=0.154\linewidth]{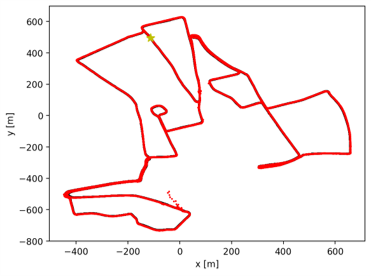} \\
    
    \raisebox{6ex}{\rotatebox[]{90}{LightLoc}} &
    \includegraphics[width=0.154\linewidth]{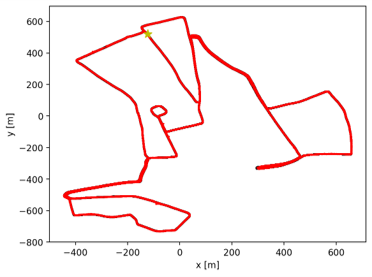} &
    \includegraphics[width=0.154\linewidth]{sec/diagram/qeoxford2639_lightloc.png} &
    \includegraphics[width=0.154\linewidth]{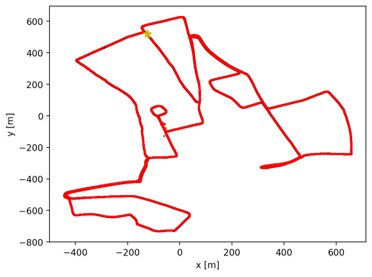} &
    \includegraphics[width=0.154\linewidth]{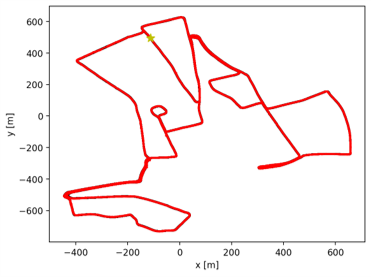} \\
    
    \raisebox{6ex}{\rotatebox[]{90}{LEADER}} &
    \includegraphics[width=0.154\linewidth]{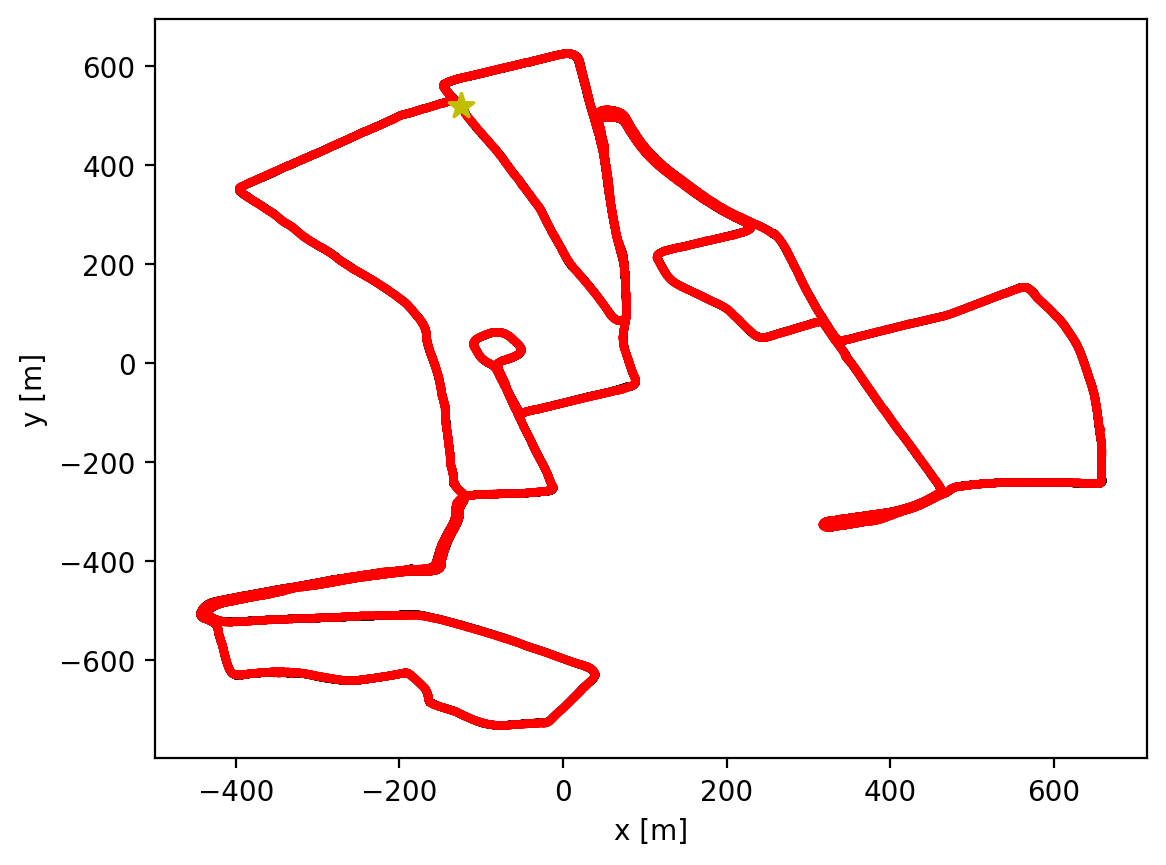} &
    \includegraphics[width=0.154\linewidth]{sec/diagram/qeoxford2639_perfectloc.png} &
    \includegraphics[width=0.154\linewidth]{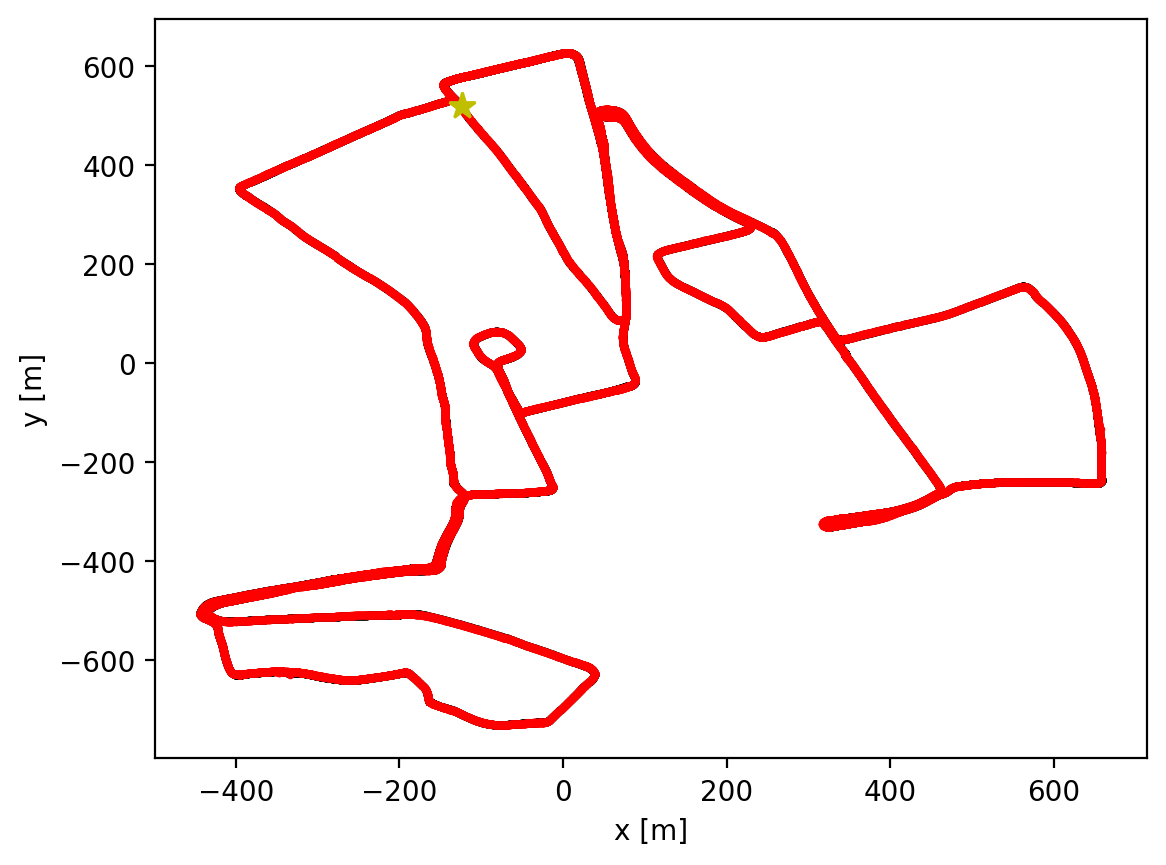} &
    \includegraphics[width=0.154\linewidth]{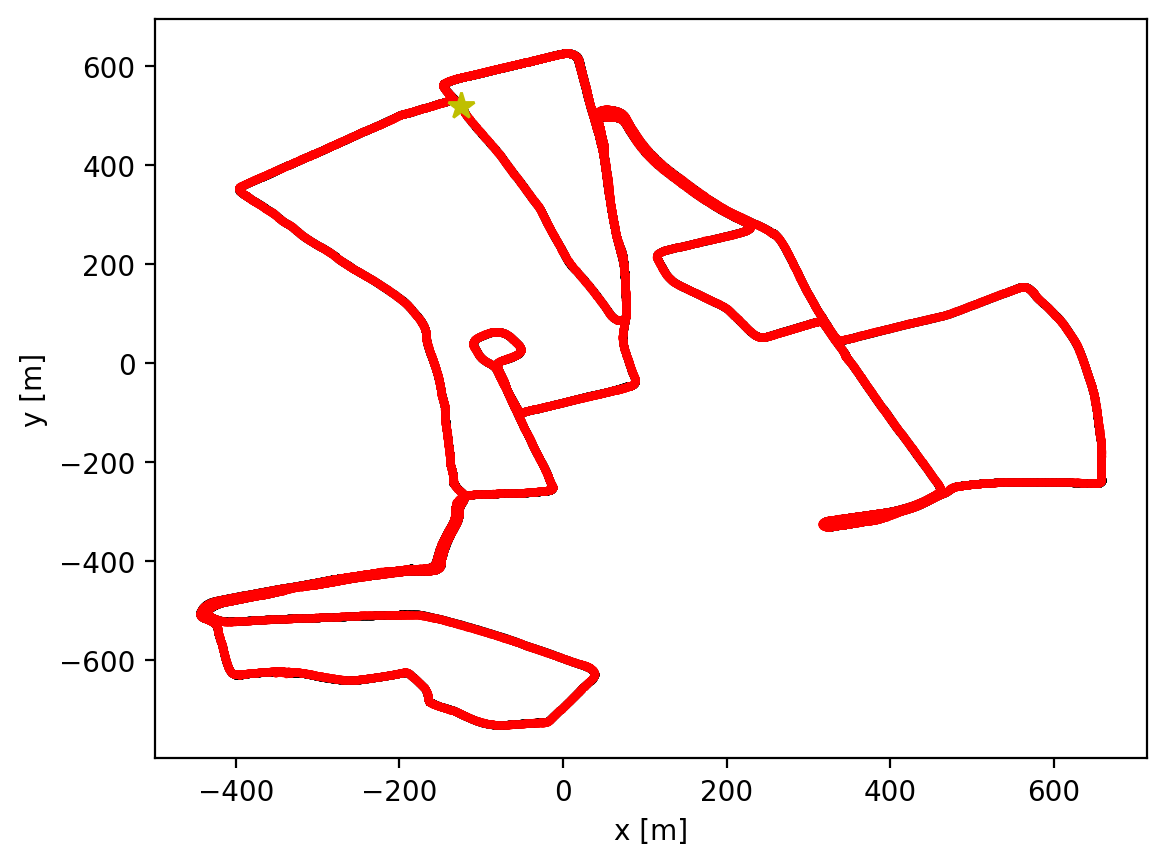} \\
\end{tabularx}

% \vspace{-5pt}
\caption{Visualization on Quality-enhanced Oxford Dataset, the star indicates the starting point.}
\label{fig:vis_nclt}

\end{table*}

\begin{table*}
\centering
\small
\setlength{\tabcolsep}{15pt}
\begin{tabularx}{\textwidth}{rcccc}
    & 2012-02-12 & 2012-02-19 & 2012-03-31 & 2012-05-26 \\
    
    \raisebox{6ex}{\rotatebox[]{90}{PNVLAD}} &
    \includegraphics[width=0.154\linewidth]{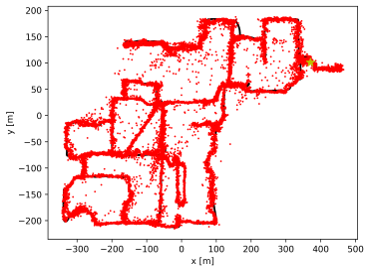} &
    \includegraphics[width=0.154\linewidth]{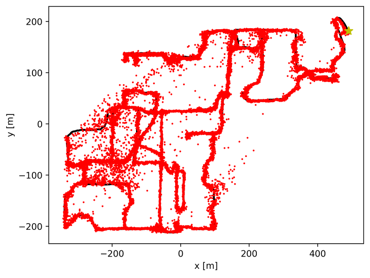} &
    \includegraphics[width=0.154\linewidth]{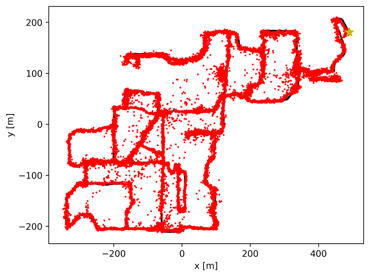} &
    \includegraphics[width=0.154\linewidth]{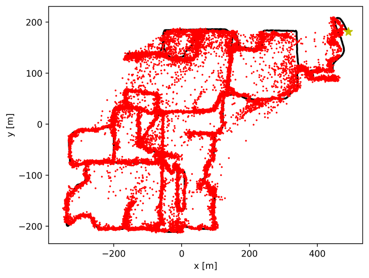} \\
    
    \raisebox{6ex}{\rotatebox[]{90}{DCP}} &
    \includegraphics[width=0.154\linewidth]{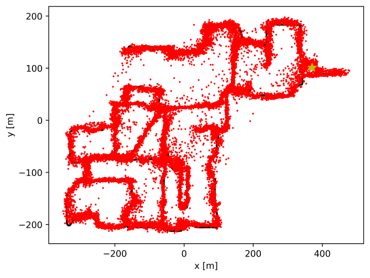} &
    \includegraphics[width=0.154\linewidth]{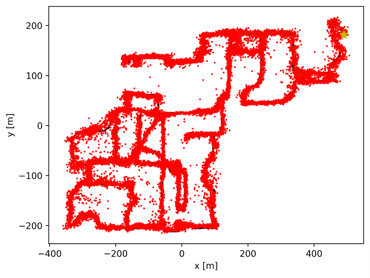} &
    \includegraphics[width=0.154\linewidth]{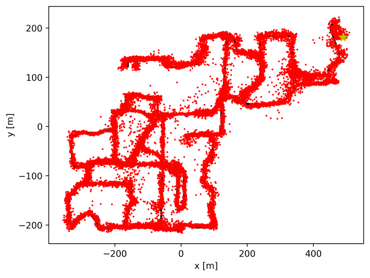} &
    \includegraphics[width=0.154\linewidth]{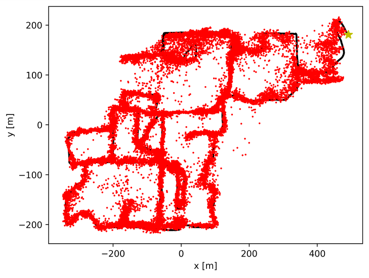} \\
    
    % \raisebox{6ex}{\rotatebox[]{90}{PosePN}} &
    % \includegraphics[width=0.154\linewidth]{sec/diagram/nclt0212_posepn.png} &
    % \includegraphics[width=0.154\linewidth]{sec/diagram/nclt0219_posepn.png} &
    % \includegraphics[width=0.154\linewidth]{sec/diagram/nclt0331_posepn.png} &
    % \includegraphics[width=0.154\linewidth]{sec/diagram/nclt0526_posepn.png} \\
    
    \raisebox{6ex}{\rotatebox[]{90}{PosePN++}} &
    \includegraphics[width=0.154\linewidth]{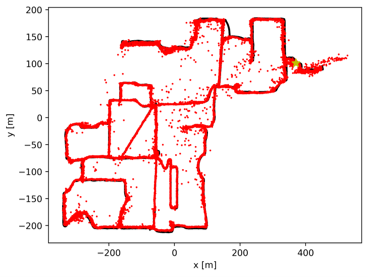} &
    \includegraphics[width=0.154\linewidth]{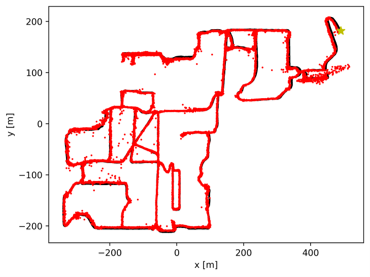} &
    \includegraphics[width=0.154\linewidth]{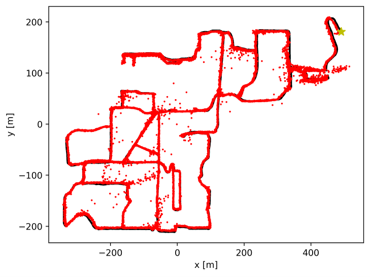} &
    \includegraphics[width=0.154\linewidth]{sec/diagram/nclt0526_posepnpp.png} \\
    
    \raisebox{6ex}{\rotatebox[]{90}{PointLoc}} &
    \includegraphics[width=0.154\linewidth]{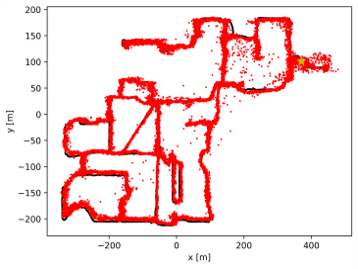} &
    \includegraphics[width=0.154\linewidth]{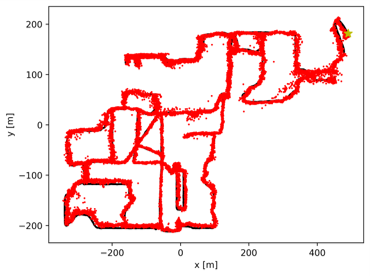} &
    \includegraphics[width=0.154\linewidth]{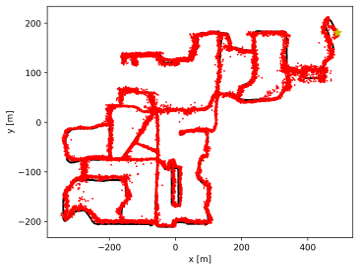} &
    \includegraphics[width=0.154\linewidth]{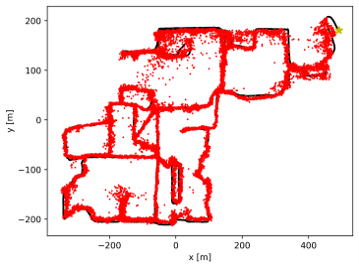} \\
    
    \raisebox{6ex}{\rotatebox[]{90}{HypLiLoc}} &
    \includegraphics[width=0.154\linewidth]{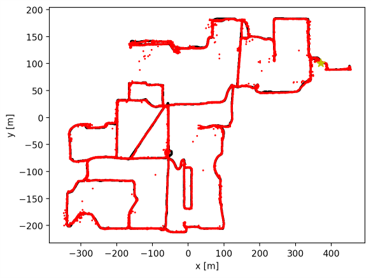} &
    \includegraphics[width=0.154\linewidth]{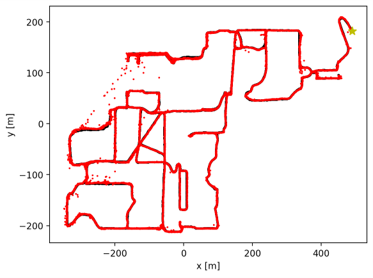} &
    \includegraphics[width=0.154\linewidth]{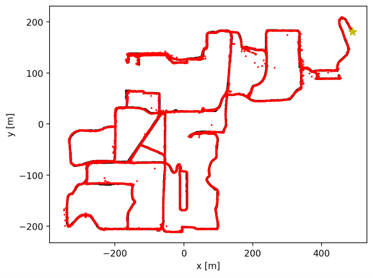} &
    \includegraphics[width=0.154\linewidth]{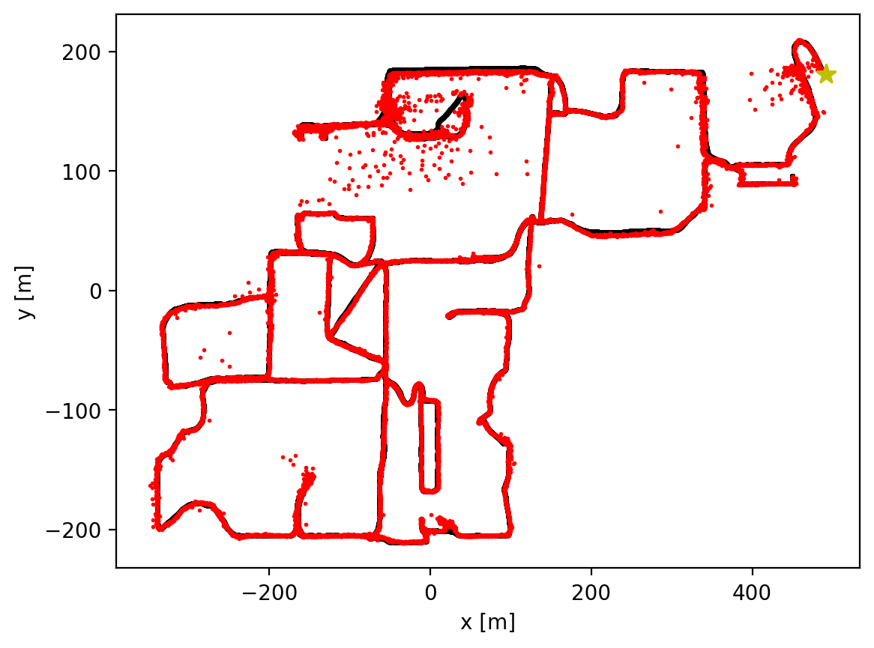} \\
    
    \raisebox{6ex}{\rotatebox[]{90}{SGLoc}} &
    \includegraphics[width=0.154\linewidth]{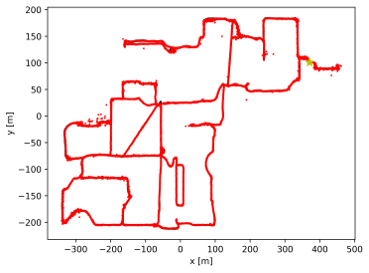} &
    \includegraphics[width=0.154\linewidth]{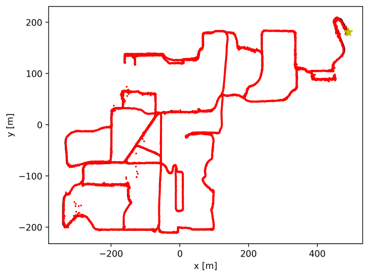} &
    \includegraphics[width=0.154\linewidth]{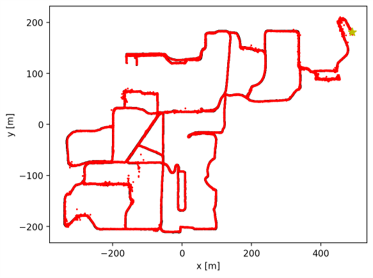} &
    \includegraphics[width=0.154\linewidth]{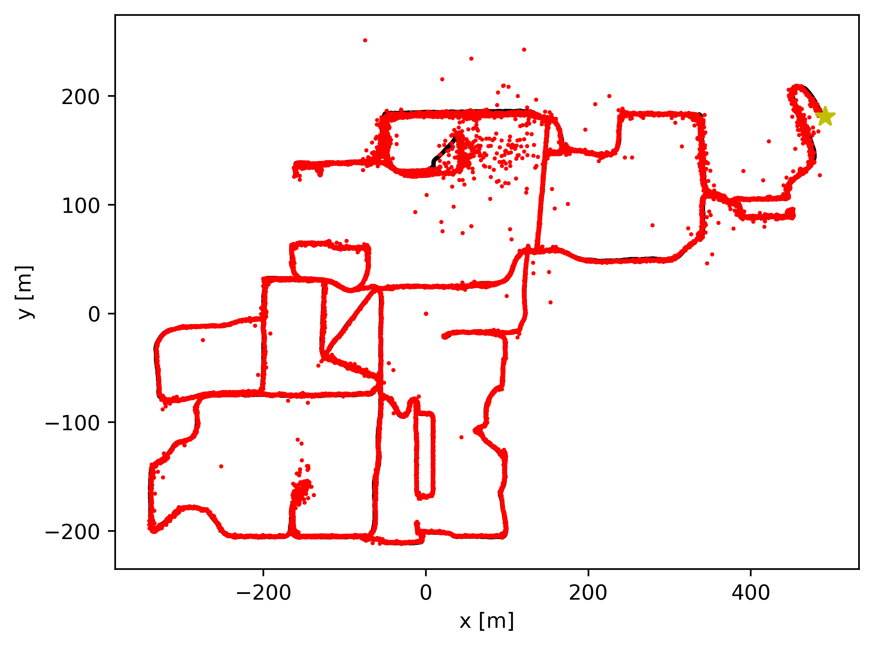} \\
    
    \raisebox{6ex}{\rotatebox[]{90}{LiSA}} &
    \includegraphics[width=0.154\linewidth]{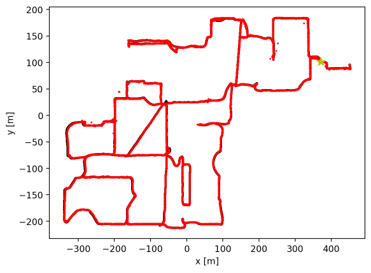} &
    \includegraphics[width=0.154\linewidth]{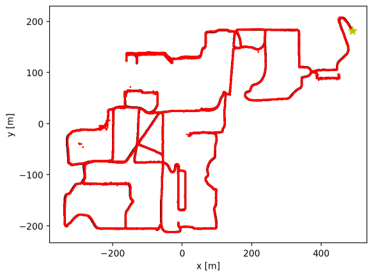} &
    \includegraphics[width=0.154\linewidth]{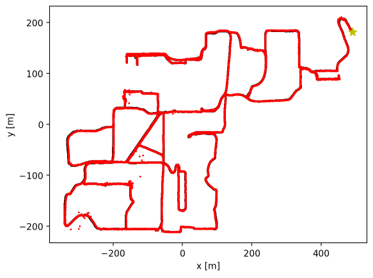} &
    \includegraphics[width=0.154\linewidth]{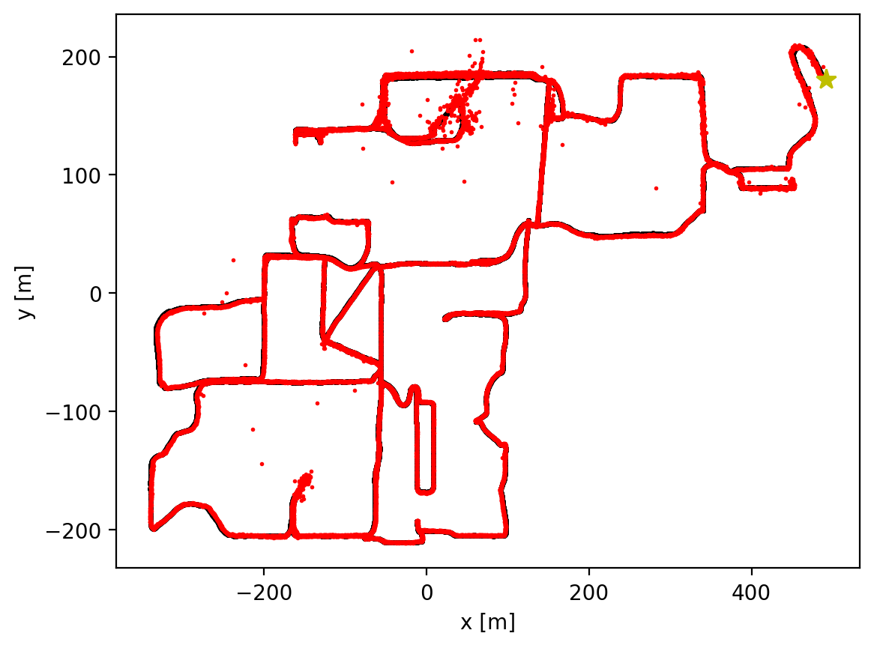} \\
    
    \raisebox{6ex}{\rotatebox[]{90}{DiffLoc}} &
    \includegraphics[width=0.154\linewidth]{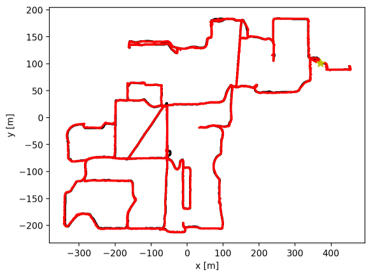} &
    \includegraphics[width=0.154\linewidth]{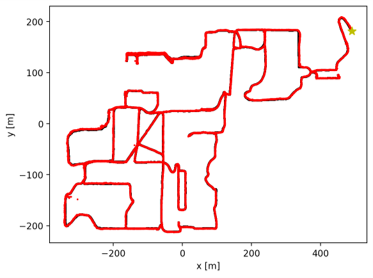} &
    \includegraphics[width=0.154\linewidth]{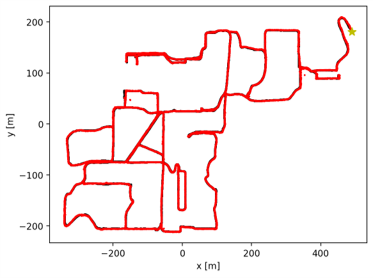} &
    \includegraphics[width=0.154\linewidth]{sec/diagram/nclt0526_diffloc.png} \\
    
    \raisebox{6ex}{\rotatebox[]{90}{LightLoc}} &
    \includegraphics[width=0.154\linewidth]{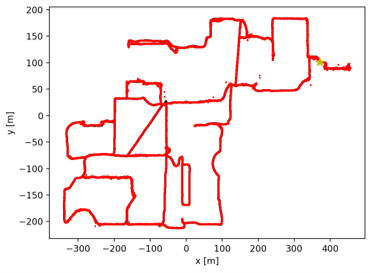} &
    \includegraphics[width=0.154\linewidth]{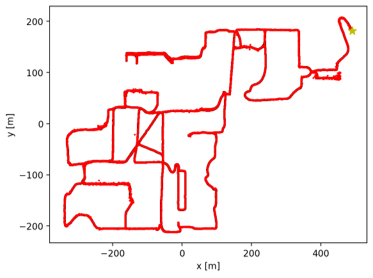} &
    \includegraphics[width=0.154\linewidth]{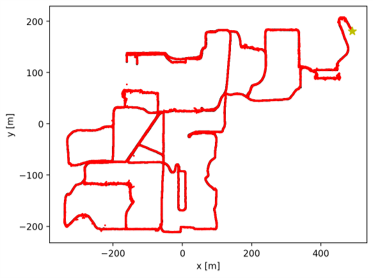} &
    \includegraphics[width=0.154\linewidth]{sec/diagram/nclt0526_lightloc.png} \\
    
    \raisebox{6ex}{\rotatebox[]{90}{LEADER}} &
    \includegraphics[width=0.154\linewidth]{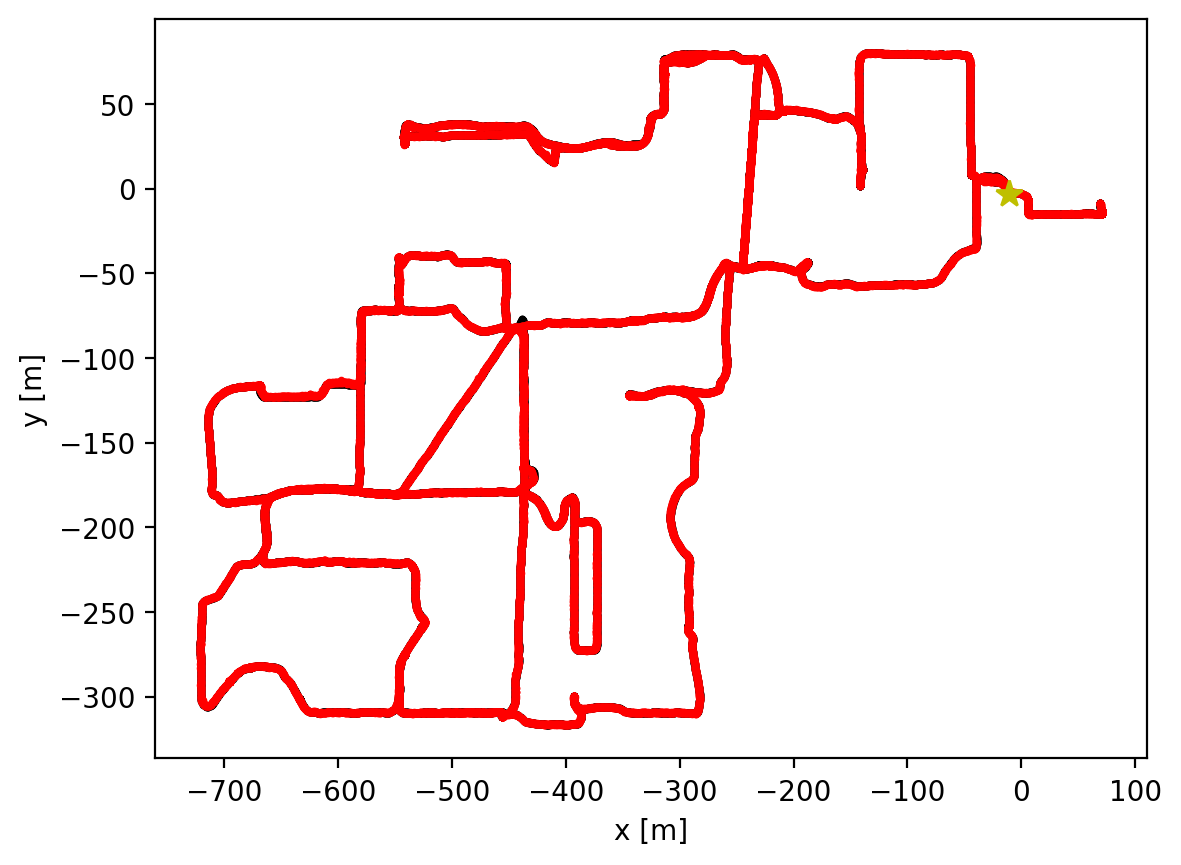} &
    \includegraphics[width=0.154\linewidth]{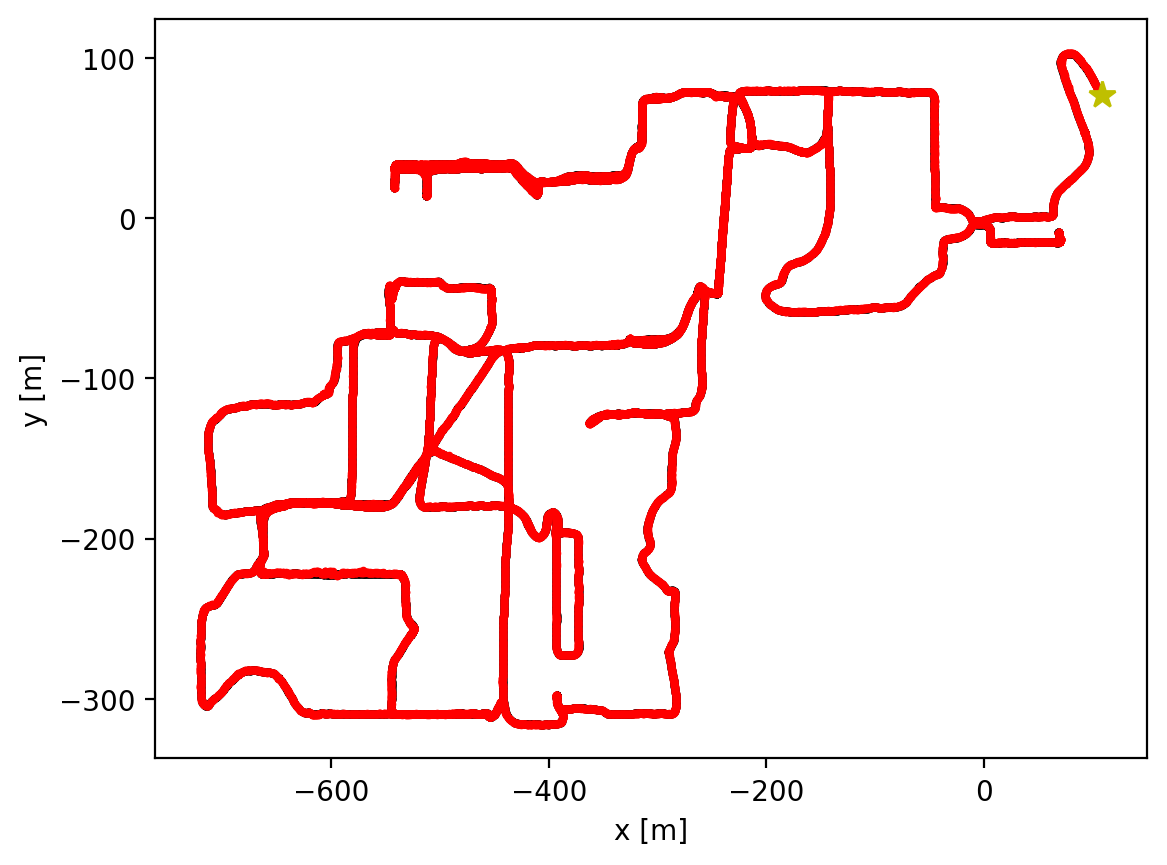} &
    \includegraphics[width=0.154\linewidth]{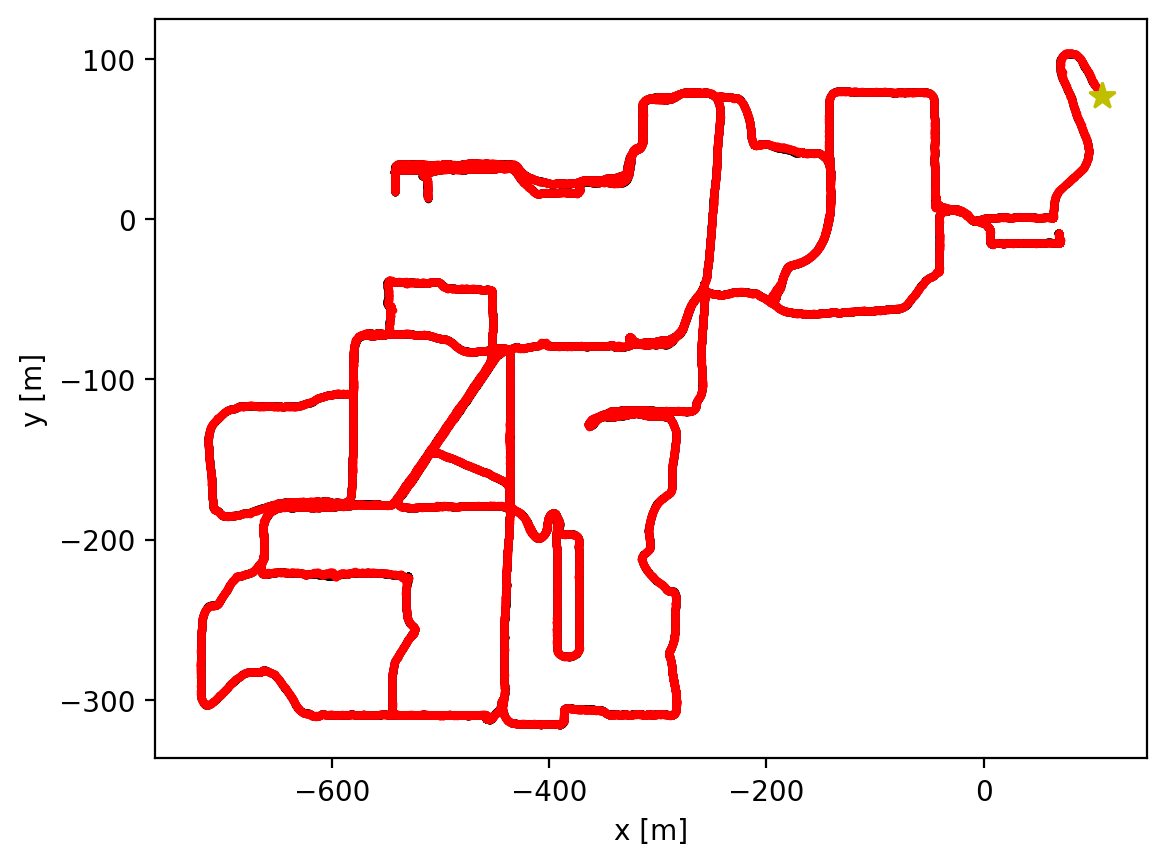} &
    \includegraphics[width=0.154\linewidth]{sec/diagram/nclt0526_perfectloc.png} \\
\end{tabularx}

\caption{Visualzation on NCLT Dataset, the star indicates the starting point.}
\label{fig:vis_oxford}
\end{table*}